\def\year{2020}\relax
\documentclass[letterpaper]{article} 
\usepackage{aaai20}  
\usepackage{times}  
\usepackage{helvet} 
\usepackage{courier}  
\usepackage[hyphens]{url}  
\usepackage{graphicx} 
\usepackage{graphicx}  
\usepackage{amssymb}
\usepackage{amsmath}
\usepackage{bm}
\usepackage{booktabs}
\usepackage[toc,page,titletoc]{appendix}

\newcommand*\samethanks[1][\value{footnote}]{\footnotemark[#1]}
\newcommand{\citet}[1]{\citeauthor{#1} \shortcite{#1}}
\newcommand{\citep}{\cite}

\title{MarioNETte: Few-shot Face Reenactment Preserving Identity of Unseen Targets}

\author{Sungjoo Ha\thanks{Equal contributions, listed in alphabetical order.}, Martin Kersner\samethanks[1], Beomsu Kim\samethanks[1], Seokjun Seo\samethanks[1], Dongyoung Kim\thanks{Corresponding author.}\\
Hyperconnect\\ 
Seoul, Republic of Korea\\
\{shurain, martin.kersner, beomsu.kim, seokjun.seo, dongyoung.kim\}@hpcnt.com}

\begin{document}
\maketitle
\begin{abstract}

When there is a mismatch between the target identity and the driver identity, face reenactment suffers severe degradation in the quality of the result, especially in a few-shot setting.
The identity preservation problem, where the model loses the detailed information of the target leading to a defective output, is the most common failure mode.
The problem has several potential sources such as the identity of the driver leaking due to the identity mismatch, or dealing with unseen large poses.
To overcome such problems, we introduce components that address the mentioned problem: image attention block, target feature alignment, and landmark transformer.
Through attending and warping the relevant features, the proposed architecture, called MarioNETte, produces high-quality reenactments of unseen identities in a few-shot setting.
In addition, the landmark transformer dramatically alleviates the identity preservation problem by isolating the expression geometry through landmark disentanglement.
Comprehensive experiments are performed to verify that the proposed framework can generate highly realistic faces, outperforming all other baselines, even under a significant mismatch of facial characteristics between the target and the driver.

\end{abstract}

\section{Introduction}

\begin{figure}[t!]
    \centering
    \includegraphics[width=1.0\columnwidth]{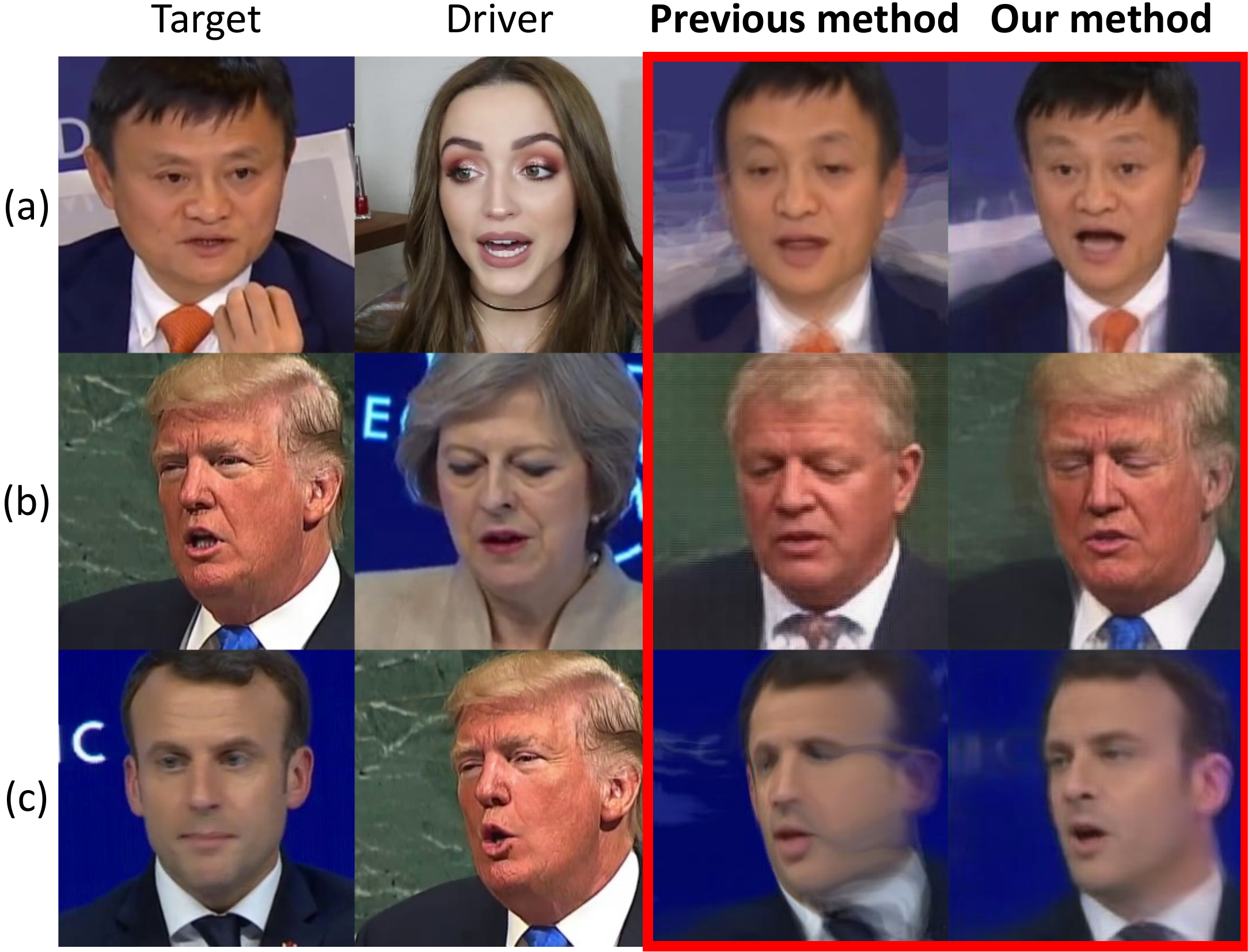}
    \caption{
    Examples of identity preservation failures and improved results generated by the proposed method. Each row shows (a) driver shape interference, (b) losing details of target identity, and (c) failure of warping at large poses.
    }
    \label{fig:problem_of_previous_few_shot}
 \end{figure}

\begin{figure*}[t]
    \includegraphics[width=0.9\linewidth]{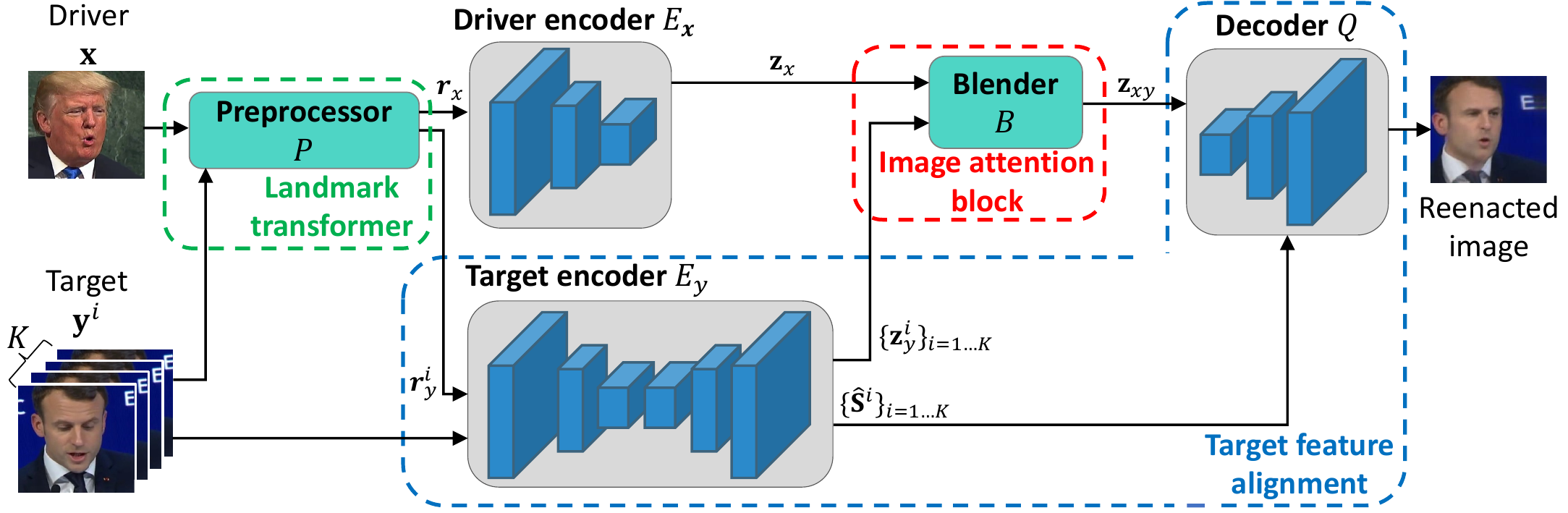}
    \centering
    \caption{
    The overall architecture of MarioNETte.
    }
    \label{fig:marionette_architecture}
 \end{figure*}

Given a \textit{target} face and a \textit{driver} face, face reenactment aims to synthesize a \textit{reenacted} face which is animated by the movement of a driver while preserving the identity of the target.

Many approaches make use of generative adversarial networks (GAN) which have demonstrated a great success in image generation tasks.
\citet{xu-arxiv-2017-facetransfer,wu-eccv-2018-reenactgan} achieved high-fidelity face reenactment results by exploiting CycleGAN~\cite{zhu-cvpr-2017-cyclegan}.
However, the CycleGAN-based approaches require at least a few minutes of training data for each target and can only reenact predefined identities, which is less attractive in-the-wild where a reenactment of unseen targets cannot be avoided.

The few-shot face reenactment approaches, therefore, try to reenact any unseen targets by utilizing operations such as adaptive instance normalization (AdaIN)~\cite{zakharov-arxiv-2019-samsung} or warping module~\cite{wiles-eccv-2018-x2face,siarohin-cvpr-2019-monkeynet}.
However, current state-of-the-art methods suffer from the problem we call \textit{identity preservation problem}: the inability to preserve the identity of the target leading to defective reenactments.
As the identity of the driver diverges from that of the target, the problem is exacerbated even further.

Examples of flawed and successful face reenactments, generated by previous approaches and the proposed model, respectively, are illustrated in Figure~\ref{fig:problem_of_previous_few_shot}.
The failures of previous approaches, for the most part, can be broken down into three different modes~\footnote{Additional example images and videos can be found at the following URL: \url{http://hyperconnect.github.io/MarioNETte}}:

\begin{enumerate}
    \item Neglecting the identity mismatch may lead to a identity of the driver interfere with the face synthesis such that the generated face resembles the driver (Figure~\ref{fig:problem_of_previous_few_shot}a).
    \item Insufficient capacity of the compressed vector representation (e.g., AdaIN layer) to preserve the information of the target identity may lead the produced face to lose the detailed characteristics (Figure~\ref{fig:problem_of_previous_few_shot}b).
    \item Warping operation incurs a defect when dealing with large poses (Figure~\ref{fig:problem_of_previous_few_shot}c).
\end{enumerate}

We propose a framework called \textit{MarioNETte}, which aims to reenact the face of unseen targets in a few-shot manner while preserving the identity without any fine-tuning.
We adopt \textit{image attention block} and \textit{target feature alignment}, which allow MarioNETte to directly inject features from the target when generating image.
In addition, we propose a novel \textit{landmark transformer} which further mitigates the identity preservation problem by adjusting for the identity mismatch in an unsupervised fashion.
Our contributions are as follows:

\begin{itemize}
    \item We propose a few-shot face reenactment framework called MarioNETte, which preserves the target identity even in situations where the facial characteristics of the driver differs widely from those of the target. Utilizing image attention block, which allows the model to attend to relevant positions of the target feature map, together with target feature alignment, which includes multiple feature-level warping operations, proposed method improves the quality of the face reenactment under different identities.
    \item We introduce a novel method of landmark transformation which copes with varying facial characteristics of different people. The proposed method adapts the landmark of a driver to that of the target in an unsupervised manner, thereby mitigating the identity preservation problem without any additional labeled data.
    \item We compare the state-of-the-art methods when the target and the driver identities coincide and differ using VoxCeleb1~\cite{nagrani-interspeech-2017-voxceleb} and CelebV~\cite{wu-eccv-2018-reenactgan} dataset, respectively. Our experiments including user studies show that the proposed method outperforms the state-of-the-art methods.
\end{itemize}

\section{MarioNETte Architecture}

Figure~\ref{fig:marionette_architecture} illustrates the overall architecture of the proposed model.
A conditional \textit{generator} $G$ generates the reenacted face given the driver $\mathbf{x}$ and the target images $\{\mathbf{y}^i\}_{i=1 \ldots K}$, and the \textit{discriminator} $D$ predicts whether the image is real or not.
The generator consists of following components:

\begin{itemize}
    \item The \textbf{preprocessor} $P$ utilizes a 3D landmark detector~\cite{bulat-iccv-2017-facealignment} to extract facial keypoints and renders them to landmark image, yielding $\mathbf{r}_x = P(\mathbf{x})$ and $\mathbf{r}_{y}^{i} = P(\mathbf{y}^{i})$, corresponding to the driver and the target input respectively. Note that proposed landmark transformer is included in the preprocessor. Since we normalize the scale, translation and rotation of landmarks before using them in a landmark transformer, we utilize 3D landmarks instead of 2D ones.

    \item The \textbf{driver encoder} $E_x(\mathbf{r}_x)$ extracts pose and expression information from the driver input and produces driver feature map $\mathbf{z}_x$.
    
    \item The \textbf{target encoder} $E_y(\mathbf{y}, \mathbf{r}_y)$ adopts a U-Net architecture to extract style information from the target input and generates target feature map $\mathbf{z}_y$ along with the warped target feature maps $\hat{\mathbf{S}}$.

    \item The \textbf{blender} $B(\mathbf{z}_x, \{\mathbf{z}_{y}^{i}\}_{i=1 \ldots K})$ receives driver feature map $\mathbf{z}_x$ and target feature maps $\mathbf{Z}_y = [\mathbf{z}_y^{1}, \ldots ,\mathbf{z}_y^{K}]$ to produce mixed feature map $\mathbf{z}_{xy}$. Proposed image attention block is basic building block of the blender.

    \item The \textbf{decoder} $Q(\mathbf{z}_{xy}, \{\hat{\mathbf{S}}^{i}\}_{i=1 \ldots K})$ utilizes warped target feature maps $\hat{\mathbf{S}}$ and mixed feature map $\mathbf{z}_{xy}$ to synthesize reenacted image. The decoder improves quality of reenacted image exploiting proposed target feature alignment.
\end{itemize}

\noindent For further details, refer to Supplementary Material A1.

\subsection{Image attention block}

\begin{figure}[t]
    \includegraphics[width=1.0\columnwidth]{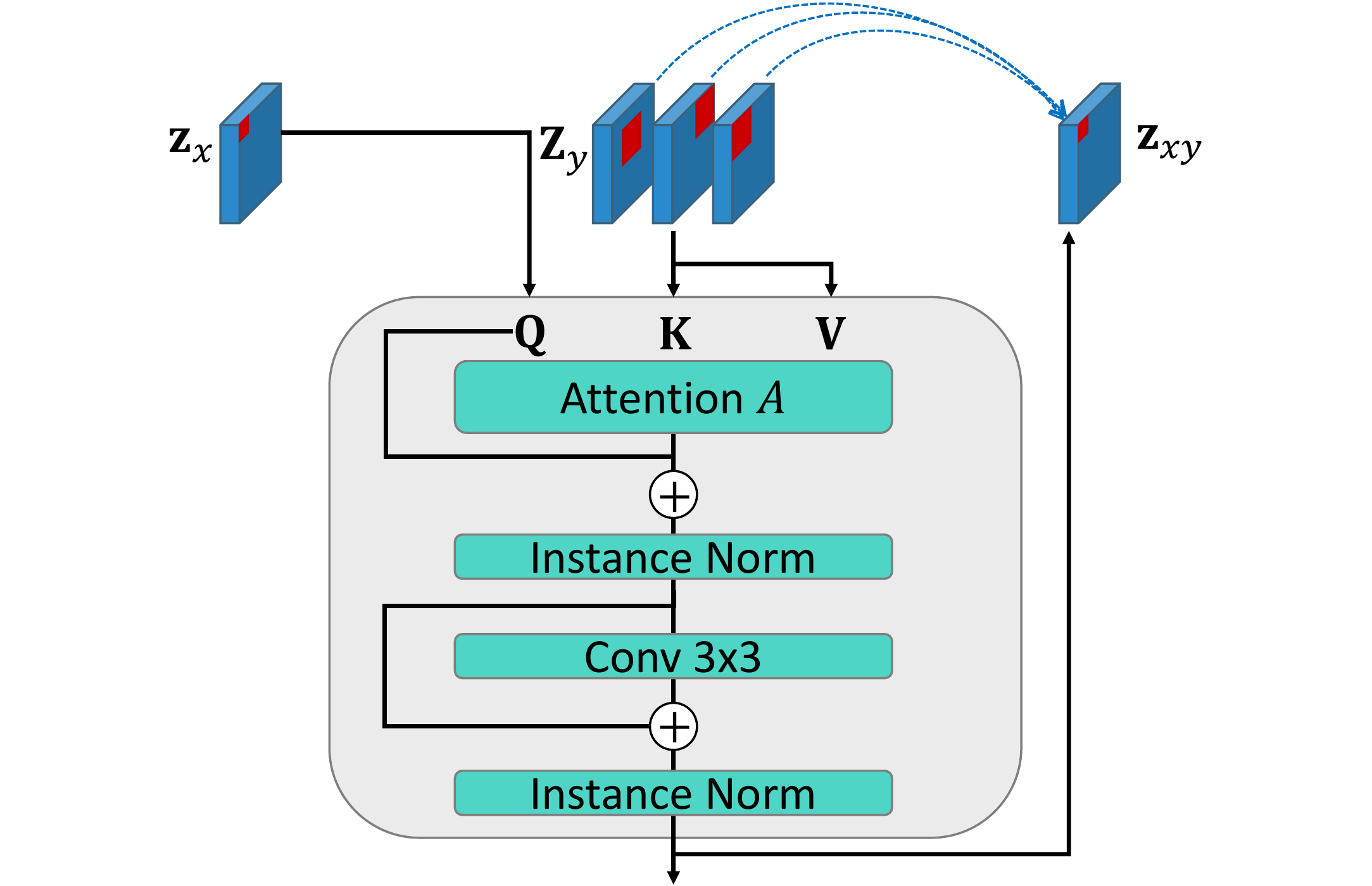}
    \caption{
    Architecture of the image attention block.
    Red boxes conceptually visualize how each position of $\mathbf{z}_x$ and $\mathbf{Z}_y$ are associated. Our attention can attend different position of each target feature maps with different importance.
    }
    \label{fig:image_attention_block}
 \end{figure}

To transfer style information of targets to the driver, previous studies encoded target information as a vector and mixed it with driver feature by concatenation or AdaIN layers~\cite{liu-arxiv-2019-funit,zakharov-arxiv-2019-samsung}.
However, encoding targets as a spatial-agnostic vector leads to losing spatial information of targets.
In addition, these methods are absent of innate design for multiple target images, and thus, summary statistics (e.g. mean or max) are used to deal with multiple targets which might cause losing details of the target.

We suggest image attention block (Figure~\ref{fig:image_attention_block}) to alleviate aforementioned problem.
The proposed attention block is inspired by the encoder-decoder attention of transformer~\cite{vaswani-nips-2017-transformer}, where the driver feature map acts as an attention query and the target feature maps act as attention memory.
The proposed attention block attends to proper positions of each feature (red boxes in Figure~~\ref{fig:image_attention_block}) while handling multiple target feature maps (i.e., $\mathbf{Z}_y$).

Given driver feature map $\mathbf{z}_x \in \mathbb{R}^{h_x \times w_x \times c_x}$ and target feature maps $\mathbf{Z}_y = [\mathbf{z}_y^{1},\ldots,\mathbf{z}_y^{K}] \in \mathbb{R}^{K \times h_y \times w_y \times c_y}$, the attention is calculated as follows:
\begin{equation}
\begin{aligned}
 \mathbf{Q} &= \mathbf{z}_{x} \mathbf{W}_{q} + \mathbf{P}_{x} \mathbf{W}_{qp} & \in &\; \mathbb{R}^{h_x \times w_x \times c_a} \\
 \mathbf{K} &= \mathbf{Z}_{y} \mathbf{W}_{k} + \mathbf{P}_{y} \mathbf{W}_{kp} & \in &\; \mathbb{R}^{K \times h_y \times w_y \times c_a} \\
 \mathbf{V} &= \mathbf{Z}_{y} \mathbf{W}_{v} & \in &\; \mathbb{R}^{K \times h_y \times w_y \times c_x} \\
\end{aligned}
\end{equation}

\begin{equation}
     A(\mathbf{Q}, \mathbf{K}, \mathbf{V}) = \text{softmax}\left(\frac{f(\mathbf{Q}) f(\mathbf{K})^T}{\sqrt{c_a}}\right)f(\mathbf{V}),
\end{equation}

\noindent where $f: \mathbb{R}^{d_1 \times \ldots \times d_k \times c} \xrightarrow{} \mathbb{R}^{(d_1 \times \ldots \times d_k) \times c}$ is a flattening function, all $\mathbf{W}$ are linear projection matrices that map to proper number of channels at the last dimension, and $\mathbf{P}_x$ and $\mathbf{P}_y$ are sinusoidal positional encodings which encode the coordinate of feature maps (further details of sinusoidal positional encodings we used are described in Supplementary Material A2).
Finally, the output $A(\mathbf{Q}, \mathbf{K}, \mathbf{V}) \in \mathbb{R}^{(h_x \times w_x) \times c_x}$ is reshaped to $\mathbb{R}^{h_x \times w_x \times c_x}$.

Instance normalization, residual connection, and convolution layer follow the attention layer to generate output feature map $\mathbf{z}_{xy}$.
The image attention block offers a direct mechanism of transferring information from multiple target images to the pose of driver.

\subsection{Target feature alignment}

\begin{figure}[t]
    \includegraphics[width=1.0\columnwidth]{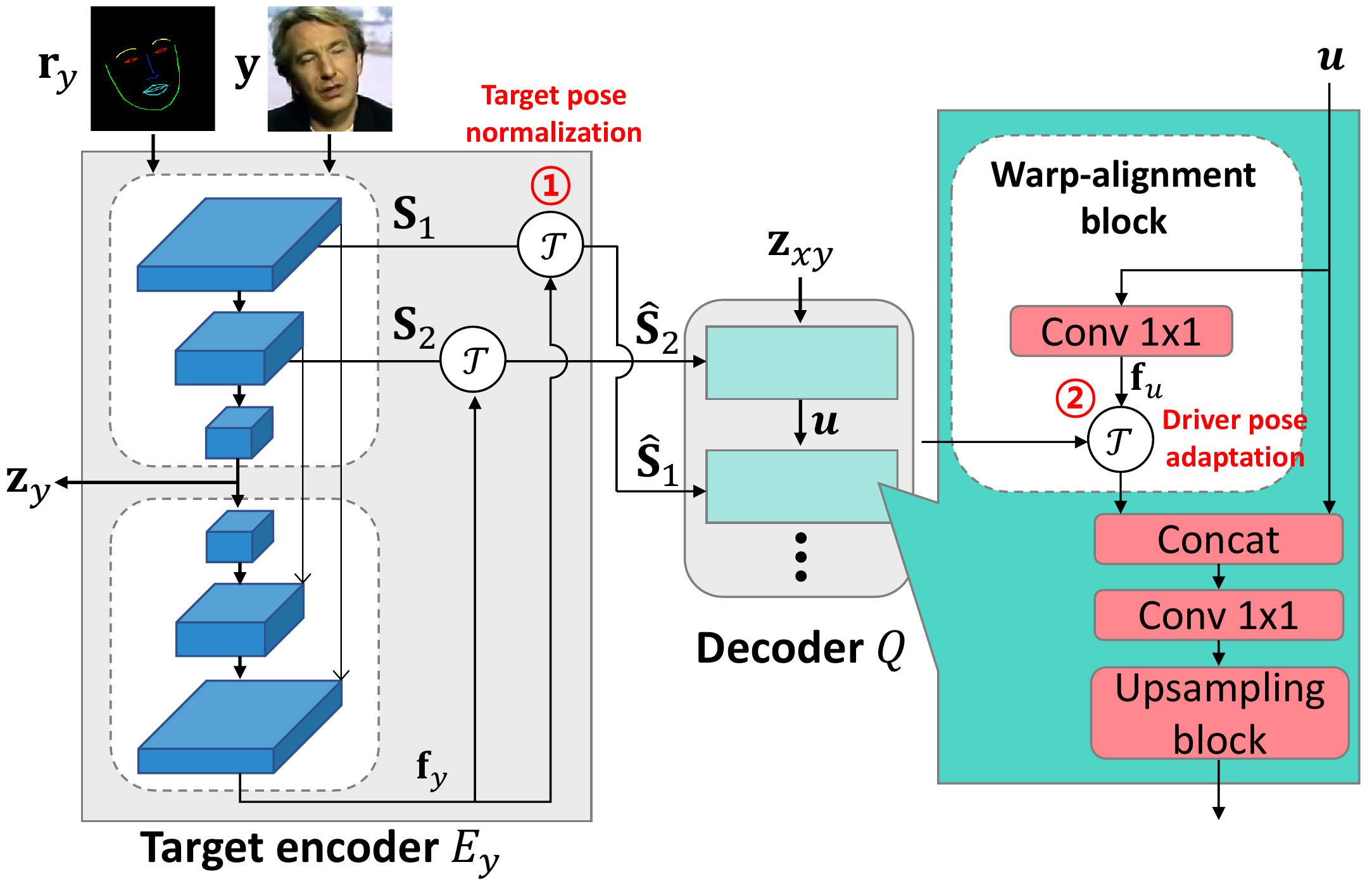}
    \caption{Architecture of target feature alignment.}
    \label{fig:style_feature_flow}
\end{figure}

The fine-grained details of the target identity can be preserved through the warping of low-level features~\cite{siarohin-cvpr-2019-monkeynet}.
Unlike previous approaches that estimate a warping flow map or an affine transform matrix by computing the difference between keypoints of the target and the driver~\cite{balakrishnan-cvpr-2018-synthesizing,siarohin-cvpr-2018-deformablegan,siarohin-cvpr-2019-monkeynet}, we propose a target feature alignment (Figure~\ref{fig:style_feature_flow}) which warps the target feature maps in two stages: (1) \textit{target pose normalization} generates pose normalized target feature maps and (2) \textit{driver pose adaptation} aligns normalized target feature maps to the pose of the driver. 
The two-stage process allows the model to better handle the structural disparities of different identities.
The details are as follows:

\begin{enumerate}
    \item \textbf{Target pose normalization.} In the target encoder $E_y$, encoded feature maps $\{\mathbf{S}_{j}\}_{j=1 \ldots n_{y}}$ are processed into $\hat{\mathbf{S}} = \{\mathcal{T}(\mathbf{S}_{1}; \mathbf{f}_{y}), \ldots ,\mathcal{T}(\mathbf{S}_{n_y}; \mathbf{f}_{y})\}$ by estimated normalization flow map $\mathbf{f}_{y}$ of target and warping function $\mathcal{T}$ (\textcircled{\raisebox{-0.9pt}{1}} in Figure~\ref{fig:style_feature_flow}).
    The following \textit{warp-alignment block} at decoder treats $\hat{\mathbf{S}}$ in a target pose-agnostic manner.

    \item \textbf{Driver pose adaptation.} The warp-alignment block in the decoder receives $\{\hat{\mathbf{S}}^i\}_{i=1 \ldots K}$ and the output $\mathbf{u}$ of the previous block of the decoder.
    In a few-shot setting, we average resolution-compatible feature maps from different target images (i.e., $\hat{\mathbf{S}}_j = \sum_i\hat{\mathbf{S}}^i_j/K$).
    To adapt pose-normalized feature maps to the pose of the driver, we generate an estimated flow map of the driver $\mathbf{f}_u$ using $1 \times 1$ convolution that takes $\mathbf{u}$ as the input.
    Alignment by $\mathcal{T}(\hat{\mathbf{S}}_j;\mathbf{f}_{u})$ follows (\textcircled{\raisebox{-0.9pt}{2}} in Figure~\ref{fig:style_feature_flow}). Then, the result is concatenated to $\mathbf{u}$ and fed into the following residual upsampling block.
\end{enumerate}

%

\section{Landmark Transformer}

Large structural differences between two facial landmarks may lead to severe degradation of the quality of the reenactment.
The usual approach to such a problem has been to learn a transformation for every identity~\cite{wu-eccv-2018-reenactgan} or by preparing a paired landmark data with the same expressions~\cite{zhang-arxiv-2019-faceswapnet}.
However, these methods are unnatural in a few-shot setting where we handle unseen identities, and moreover, the labeled data is hard to be acquired.
To overcome this difficulty, we propose a novel \textit{landmark transformer} which transfers the facial expression of the driver to an arbitrary target identity.
The landmark transformer utilizes multiple videos of unlabeled human faces and is trained in an unsupervised manner.

\subsection{Landmark decomposition}

Given video footages of different identities, we denote $\mathbf{x}(c,t)$ as the $\mathit{t}$-th frame of the $\mathit{c}$-th video, and $\mathbf{l}(c,t)$ as a 3D facial landmark.
We first transform every landmark into a normalized landmark $\mathbf{\bar{l}}(c,t)$ by normalizing the scale, translation, and rotation.
Inspired by 3D morphable models of face~\cite{blanz-siggraph-1999-3dmm}, we assume that normalized landmarks can be decomposed as follows:

\begin{equation}
\begin{aligned}
    \mathbf{\bar{l}}(c,t) &= \mathbf{\bar{l}}_m + \mathbf{\bar{l}}_{id}(c) + \mathbf{\bar{l}}_{exp}(c, t),
\end{aligned}
\label{eq:landmark_decompose}
\end{equation}
where $\mathbf{\bar{l}}_m$ is the average facial landmark geometry computed by taking the mean over all landmarks, $\mathbf{\bar{l}}_{id}(c)$ denotes the landmark geometry of identity $c$, computed by $\mathbf{\bar{l}}_{id}(c) = \sum_{t} \mathbf{\bar{l}}(c,t)/{T_c} - \mathbf{\bar{l}}_m$ where $T_c$ is the number of frames of $c$-th video, and $\mathbf{\bar{l}}_{exp}(c, t)$ corresponds to the expression geometry of $\mathit{t}$-th frame.
The decomposition leads to $\mathbf{\bar{l}}_{exp}(c,t) = \mathbf{\bar{l}}(c,t) - \mathbf{\bar{l}}_m - \mathbf{\bar{l}}_{id}(c)$.

Given a target landmark $\mathbf{\bar{l}}(c_y,t_y)$ and a driver landmark $\mathbf{\bar{l}}(c_x,t_x)$ we wish to generate the following landmark:
\begin{equation}
    \mathbf{\bar{l}}(c_x \xrightarrow{} c_y,t_x) = \mathbf{\bar{l}}_m + \mathbf{\bar{l}}_{id}(c_y) + \mathbf{\bar{l}}_{exp}(c_x, t_x),
\label{eq:landmark_transformation}
\end{equation}
i.e., a landmark with the identity of the target and the expression of the driver.
Computing $\mathbf{\bar{l}}_{id}(c_y)$ and $\mathbf{\bar{l}}_{exp}$ is possible if enough images of $c_y$ are given, but in a few-shot setting, it is difficult to disentangle landmark of unseen identity into two terms.

\subsection{Landmark disentanglement}

To decouple the identity and the expression geometry in a few-shot setting, we introduce a neural network to regress the coefficients for linear bases.
Previously, such an approach has been widely used in modeling complex face geometries~\cite{blanz-siggraph-1999-3dmm}.
We separate expression landmarks into semantic groups of the face (e.g., mouth, nose and eyes) and perform PCA on each group to extract the expression bases from the training data:
\begin{equation}
    \mathbf{\bar{l}}_{exp}(c,t) = \sum_{k=1}^{n_{exp}} \alpha_k(c,t) \mathbf{b}_{exp, k} = \mathbf{b}_{exp}^{T} \bm{\alpha}(c,t),
    \label{eq:expression_regress}
\end{equation}
where $\mathbf{b}_{exp, k}$ and $\alpha_k$ represent the basis and the corresponding coefficient, respectively.

\begin{figure}[t]
    \includegraphics[width=1.0\columnwidth]{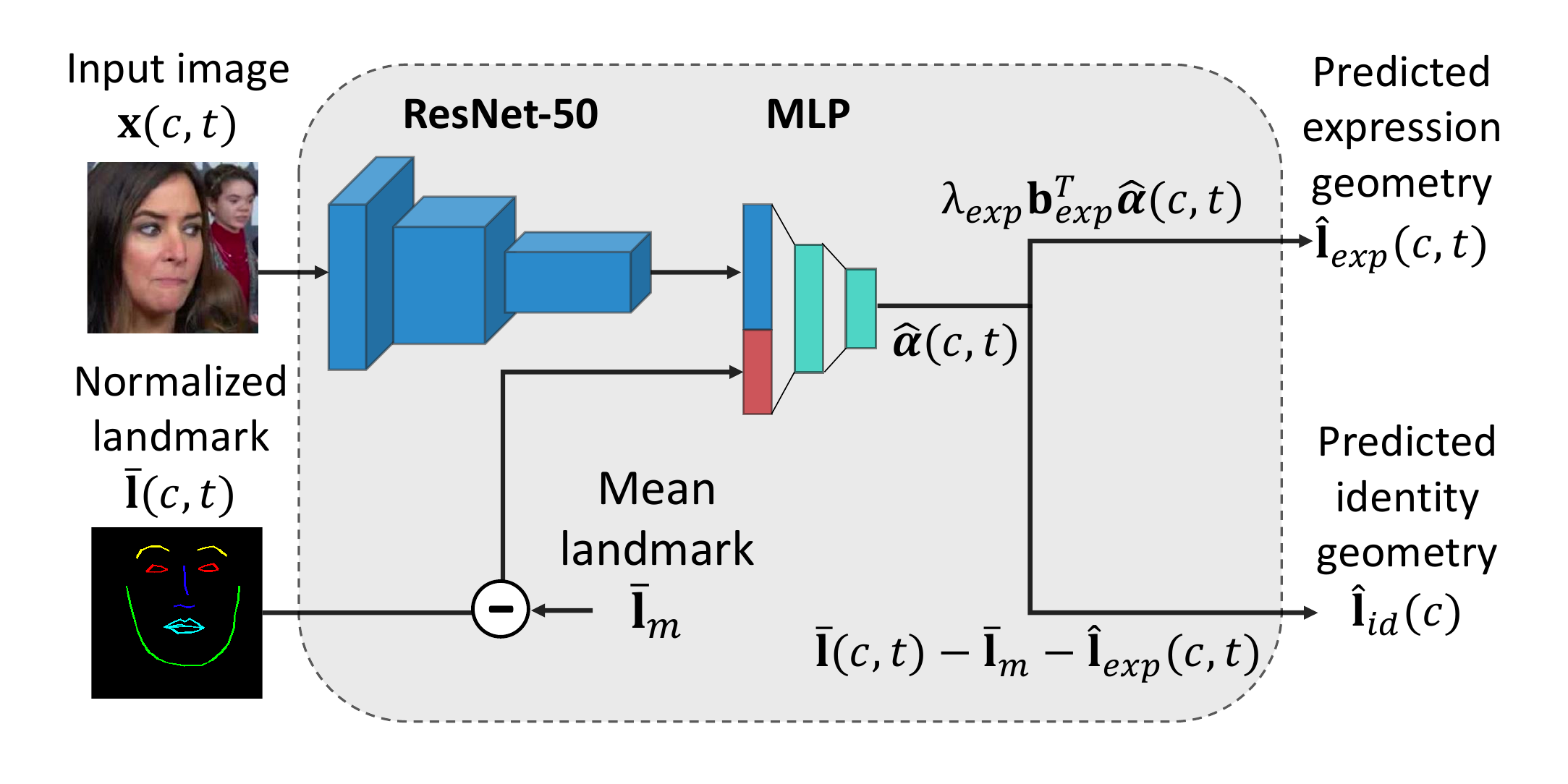}
    \caption{
    Architecture of landmark disentangler.
    Note that $\mathbf{\bar{l}}(c, t)$ is a set of landmark points but visualized as an image in the figure.
    }
    \label{fig:landmark_transform_architecture}
 \end{figure}

The proposed neural network, a \textit{landmark disentangler} $M$, estimates $\bm{\alpha}(c,t)$ given an image $\mathbf{x}(c,t)$ and a landmark $\mathbf{\bar{l}}(c,t)$.
Figure~\ref{fig:landmark_transform_architecture} illustrates the architecture of the landmark disentangler.
Once the model is trained, the identity and the expression geometry can be computed as follows:
\begin{equation}
\begin{aligned}
    \bm{\hat{\alpha}}(c,t) &= M\left(\mathbf{x}(c,t), \;  \mathbf{\bar{l}}(c,t)\right) \\
    \mathbf{\hat{l}}_{exp}(c,t) &= \lambda_{exp} \mathbf{b}_{exp}^T \bm{\hat{\alpha}}(c,t) \\
    \mathbf{\hat{l}}_{id}(c) &= \mathbf{\bar{l}}(c,t) - \mathbf{\bar{l}}_m - \mathbf{\hat{l}}_{exp}(c,t),
\end{aligned}
\label{eq:estimate_id_and_exp}
\end{equation}
where $\lambda_{exp}$ is a hyperparameter that controls the intensity of the predicted expressions from the network.
Image feature extracted by a ResNet-50 and the landmark, $\mathbf{\bar{l}}(c,t) - \mathbf{\bar{l}}_m$, are fed into a 2-layer MLP to predict $\bm{\hat{\alpha}}(c,t)$.

During the inference, the target and the driver landmarks are processed according to Equation~\ref{eq:estimate_id_and_exp}.
When multiple target images are given, we take the mean value over all $\mathbf{\hat{l}}_{id}(c_y)$.
Finally, landmark transformer converts landmark as:
\begin{equation}
    \mathbf{\hat{l}}(c_x \xrightarrow{} c_y,t_x) = \mathbf{\bar{l}}_m + \mathbf{\hat{l}}_{id}(c_y) + \mathbf{\hat{l}}_{exp}(c_x, t_x).
\end{equation}
Denormalization to recover the original scale, translation, and rotation is followed by the rasterization that generates a landmark adequate for the generator to consume.
Further details of landmark transformer are described in Supplementary Material B.

\section{Experimental Setup}

\subsubsection{Datasets}

We trained our model and the baselines using VoxCeleb1~\cite{nagrani-interspeech-2017-voxceleb}, which contains $256\times256$ size videos of 1,251 different identities.
We utilized the test split of VoxCeleb1 and CelebV~\cite{wu-eccv-2018-reenactgan} for evaluating self-reenactment and reenactment under a different identity, respectively.
We created the test set by sampling 2,083 image sets from randomly selected 100 videos of VoxCeleb1 test split, and uniformly sampled 2,000 image sets from every identity from CelebV.
The CelebV data includes the videos of five different celebrities of widely varying characteristics, which we utilize to evaluate the performance of the models reenacting unseen targets, similar to in-the-wild scenario.
Further details of the loss function and the training method can be found at Supplementary Material A3 and A4.

\begin{figure*}[t]
    \includegraphics[width=1.0\linewidth]{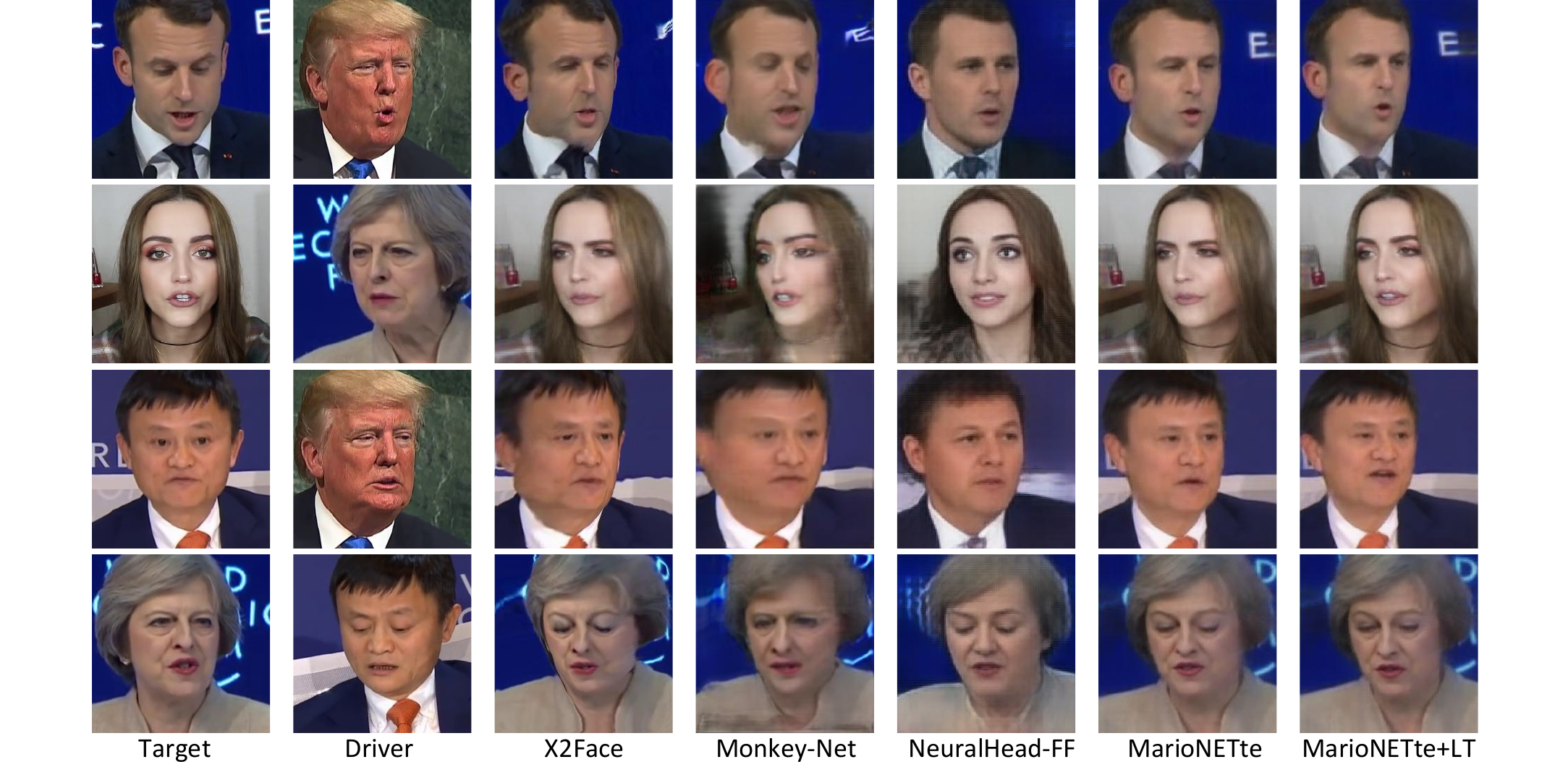}
    \centering
    \caption{Images generated by the proposed method and baselines, reenacting different identity on CelebV in one-shot setting.}
    \label{fig:main_result}
 \end{figure*}

\subsubsection{Baselines}

MarioNETte variants, with and without the landmark transformer (\textit{MarioNETte+LT} and \textit{MarioNETte}, respectively), are compared with state-of-the-art models for few-shot face reenactment.
Details of each baseline are as follows:

\begin{itemize}
    \item \textbf{X2Face}~\cite{wiles-eccv-2018-x2face}. X2face utilizes direct image warping. We used the pre-trained model provided by the authors, trained on VoxCeleb1.
    
    \item \textbf{Monkey-Net}~\cite{siarohin-cvpr-2019-monkeynet}. Monkey-Net adopts feature-level warping. We used the implementation provided by the authors. Due to the structure of the method, Monkey-Net can only receive a single target image.
    
    \item \textbf{NeuralHead}~\cite{zakharov-arxiv-2019-samsung}.
    NeuralHead exploits AdaIN layers. Since a reference implementation is absent, we made an honest attempt to reproduce the results.
    Our implementation is a feed-forward version of their model (\textit{NeuralHead-FF}) where we omit the meta-learning as well as fine-tuning phase, because we are interested in using a single model to deal with multiple identities.
\end{itemize}

\subsubsection{Metrics}

We compare the models based on the following metrics to evaluate the quality of the generated images.
Structured similarity (\textbf{SSIM})~\cite{wang-tip-2004-ssim} and peak signal-to-noise ratio (\textbf{PSNR}) evaluate the low-level similarity between the generated image and the ground-truth image.
We also report the masked-SSIM (\textbf{M-SSIM}) and masked-PSNR (\textbf{M-PSNR}) where the measurements are restricted to the facial region.

In the absence of the ground truth image where different identity drives the target face, the following metrics are more relevant.
Cosine similarity (\textbf{CSIM}) of embedding vectors generated by pre-trained face recognition model~\cite{deng-cvpr-2019-csim} is used to evaluate the quality of identity preservation.
To inspect the capability of the model to properly reenact the pose and the expression of the driver, we compute \textbf{PRMSE}, the root mean square error of the head pose angles, and \textbf{AUCON}, the ratio of identical facial action unit values, between the generated images and the driving images.
OpenFace~\cite{baltrusaitis-ieee-2018-openface} is utilized to compute pose angles and action unit values.

\section{Experimental Results}

Models were compared under self-reenactment and reenactment of different identities, including a user study.
Ablation tests were conducted as well.
All experiments were conducted under two different settings: one-shot and few-shot, where one or eight target images were used respectively.

\subsection{Self-reenactment}

\begin{table*}[t]
    \centering
    \begin{tabular}{ccccccccc}
        \toprule
Model (\# target) & CSIM$\uparrow$ & SSIM$\uparrow$ & M-SSIM$\uparrow$ & PSNR$\uparrow$ & M-PSNR$\uparrow$ & PRMSE$\downarrow$ & AUCON$\uparrow$ \\
\midrule
X2face (1) & 0.689 & 0.719 & 0.941 & 22.537 & 31.529 & 3.26 & 0.813 \\ 
Monkey-Net (1) & 0.697 & 0.734 & 0.934 & \bf{23.472} & 30.580 & 3.46 & 0.770 \\
NeuralHead-FF (1) & 0.229 & 0.635 & 0.923 & 20.818 & 29.599 & 3.76 & 0.791 \\
MarioNETte (1) & \bf{0.755} & \bf{0.744} & \bf{0.948} & 23.244 & \bf{32.380} & \bf{3.13} & \bf{0.825} \\
\midrule
X2face (8) & 0.762 & 0.776 & 0.956 & 24.326 & 33.328 & 3.21 & 0.826 \\
NeuralHead-FF (8) & 0.239 & 0.645 & 0.925 & 21.362 & 29.952 & 3.69 & 0.795 \\
MarioNETte (8) & \bf{0.828} & \bf{0.786} & \bf{0.958} & \bf{24.905} & \bf{33.645} & \bf{2.57} & \bf{0.850} \\

        \bottomrule
    \end{tabular}
    \caption{Evaluation result of self-reenactment setting on VoxCeleb1. Upward/downward pointing arrows correspond to metrics that are better when the values are higher/lower.}
    \label{tab:self_reenact_result}
\end{table*} 

Table~\ref{tab:self_reenact_result} illustrates the evaluation results of the models under self-reenactment settings on VoxCeleb1.
\textit{MarioNETte} surpasses other models in every metric under few-shot setting and outperforms other models in every metric except for PSNR under the one-shot setting.
However, \textit{MarioNETte} shows the best performance in M-PSNR which implies that it performs better on facial region compared to baselines.
The low CSIM yielded from \textit{NeuralHead-FF} is an indirect evidence of the lack of capacity in AdaIN-based methods.

\subsection{Reenacting Different Identity}

\begin{table}[t]
    \centering
    \begin{tabular}{cccc}
        \toprule
Model (\# target) & CSIM$\uparrow$ & PRMSE$\downarrow$ & AUCON$\uparrow$ \\
\midrule
X2face (1) & 0.450 & 3.62 & 0.679 \\
Monkey-Net (1) & 0.451 & 4.81 & 0.584 \\
NeuralHead-FF (1) & 0.108 & \bf{3.30} & \bf{0.722} \\
MarioNETte (1) & \underline{0.520} & \underline{3.41} & \underline{0.710} \\
MarioNETte+LT (1) & \bf{0.568} & 3.70 & 0.684 \\
\midrule
X2face (8) & 0.484 & \bf{3.15} & 0.709 \\
NeuralHead-FF (8) & 0.120 & \underline{3.26} & \bf{0.723} \\
MarioNETte (8) & \underline{0.608} & \underline{3.26} & \underline{0.717} \\
MarioNETte+LT (8) & \bf{0.661} & 3.57 & 0.691 \\
        \bottomrule
    \end{tabular}
    \caption{Evaluation result of reenacting a different identity on CelebV. \textbf{Bold} and \underline{underlined} values correspond to the best and the second-best value of each metric, respectively.}
    \label{tab:diff_reenact_result}
\end{table}

\begin{table}[ht]
    \centering
    \begin{tabular}{cccc}
        \toprule
        Model (\# target) & \begin{tabular}[c]{@{}c@{}}vs. \\Ours\end{tabular} & \begin{tabular}[c]{@{}c@{}}vs. \\Ours+LT\end{tabular} & Realism $\uparrow$\\
        \midrule
        X2Face (1) & 0.07 & 0.09 & 0.093 \\
        Monkey-Net (1) & 0.05 & 0.09 & 0.100 \\
        NeuralHead-FF (1)& 0.17 & 0.17 & 0.087 \\
        MarioNETte (1) & - & 0.51 & 0.140 \\
        MarioNETte+LT (1) & - & - & \bf{0.187} \\
        \midrule
        \midrule
        X2Face (8) & 0.09 & 0.07 & 0.047 \\
        NeuralHead-FF (8) & 0.15 & 0.16 & 0.080 \\
        MarioNETte (8)& - & 0.52 & 0.147 \\
        MarioNETte+LT (8) & - & - & \bf{0.280} \\
        \bottomrule
    \end{tabular}
    \caption{User study results of reenacting different identity on CelebV. \textit{Ours} stands for our proposed model, \textit{MarioNETte}, and \textit{Ours+LT} stands for \textit{MarioNETte+LT}.}
   
    \label{tab:user_study}
    
\end{table}

\begin{table}[ht]
    \centering
    \begin{tabular}{cccc}
        \toprule
Model (\# target) & CSIM$\uparrow$ & PRMSE$\downarrow$ & AUCON$\uparrow$ \\
\midrule
AdaIN (1) & 0.063 & 3.47 & \underline{0.724} \\
+Attention (1) & 0.333 & \bf{3.17} & \bf{0.729} \\
+Alignment (1) & \bf{0.530} & 3.44 & 0.700 \\
\midrule
MarioNETte (1) & \underline{0.520} & \underline{3.41} & 0.710 \\
\midrule
\midrule
AdaIN (8) & 0.069 & 3.40 & \underline{0.723} \\
+Attention (8) & 0.472 & \bf{3.22} & \bf{0.727} \\
+Alignment (8) & \underline{0.605} & 3.27 & 0.709 \\
\midrule
MarioNETte (8) & \bf{0.608} &\underline{3.26} & 0.717 \\
\bottomrule
    \end{tabular}
    \caption{Comparison of ablation models for reenacting different identity on CelebV.}
    \label{tab:ablation_result}
\end{table}

Table~\ref{tab:diff_reenact_result} displays the evaluation result of reenacting a different identity on CelebV, and Figure~\ref{fig:main_result} shows generated images from proposed method and baselines.
\textit{MarioNETte} and \textit{MarioNETte+LT} preserve target identity adequately, thereby outperforming other models in CSIM.
The proposed method alleviates the identity preservation problem regardless of the driver being of the same identity or not.
While \textit{NeuralHead-FF} exhibits slightly better performance in terms of PRMSE and AUCON compared to \textit{MarioNETte}, the low CSIM of \textit{NeuralHead-FF} portrays the failure to preserve the target identity.
The landmark transformer significantly boosts identity preservation at the cost of a slight decrease in PRMSE and AUCON. The decrease may be due to the PCA bases for the expression disentanglement not being diverse enough to span the whole space of expressions. Moreover, the disentanglement of identity and expression itself is a non-trivial problem, especially in a one-shot setting.

\subsection{User Study}

Two types of user studies are conducted to assess the performance of the proposed model:

\begin{itemize}
\item \textbf{Comparative analysis.} Given three example images of the target and a driver image, we displayed two images generated by different models and asked human evaluators to select an image with higher quality.
The users were asked to assess the quality of an image in terms of (1) identity preservation, (2) reenactment of driver's pose and expression, and (3) photo-realism.
We report the winning ratio of baseline models compared to our proposed models.
We believe that user reported score better reflects the quality of different models than other indirect metrics.

\item \textbf{Realism analysis.} Similar to the user study protocol of \citet{zakharov-arxiv-2019-samsung}, three images of the same person, where two of the photos were taken from a video and the remaining generated by the model, were presented to human evaluators.
Users were instructed to choose an image that differs from the other two in terms of the identity under a three-second time limit.
We report the ratio of deception, which demonstrates the identity preservation and the photo-realism of each model.
\end{itemize}

\noindent For both studies, 150 examples were sampled from CelebV, which were evenly distributed to 100 different human evaluators.

Table~\ref{tab:user_study} illustrates that our models are preferred over existing methods achieving realism scores with a large margin.
The result demonstrates the capability of \textit{MarioNETte} in creating photo-realistic reenactments while preserving the target identity in terms of human perception.
We see a slight preference of \textit{MarioNETte} over \textit{MarioNETte+LT}, which agrees with the Table~\ref{tab:diff_reenact_result}, as \textit{MarioNETte+LT} has better identity preservation capability at the expense of slight degradation in expression transfer.
Since the identity preservation capability of \textit{MarioNETte+LT} surpasses all other models in realism score, almost twice the score of even \textit{MarioNETte} on few-shot settings, we consider the minor decline in expression transfer a good compromise.

\subsection{Ablation Test}

\begin{figure}[t]
    \centering
    \includegraphics[width=1.0\columnwidth]{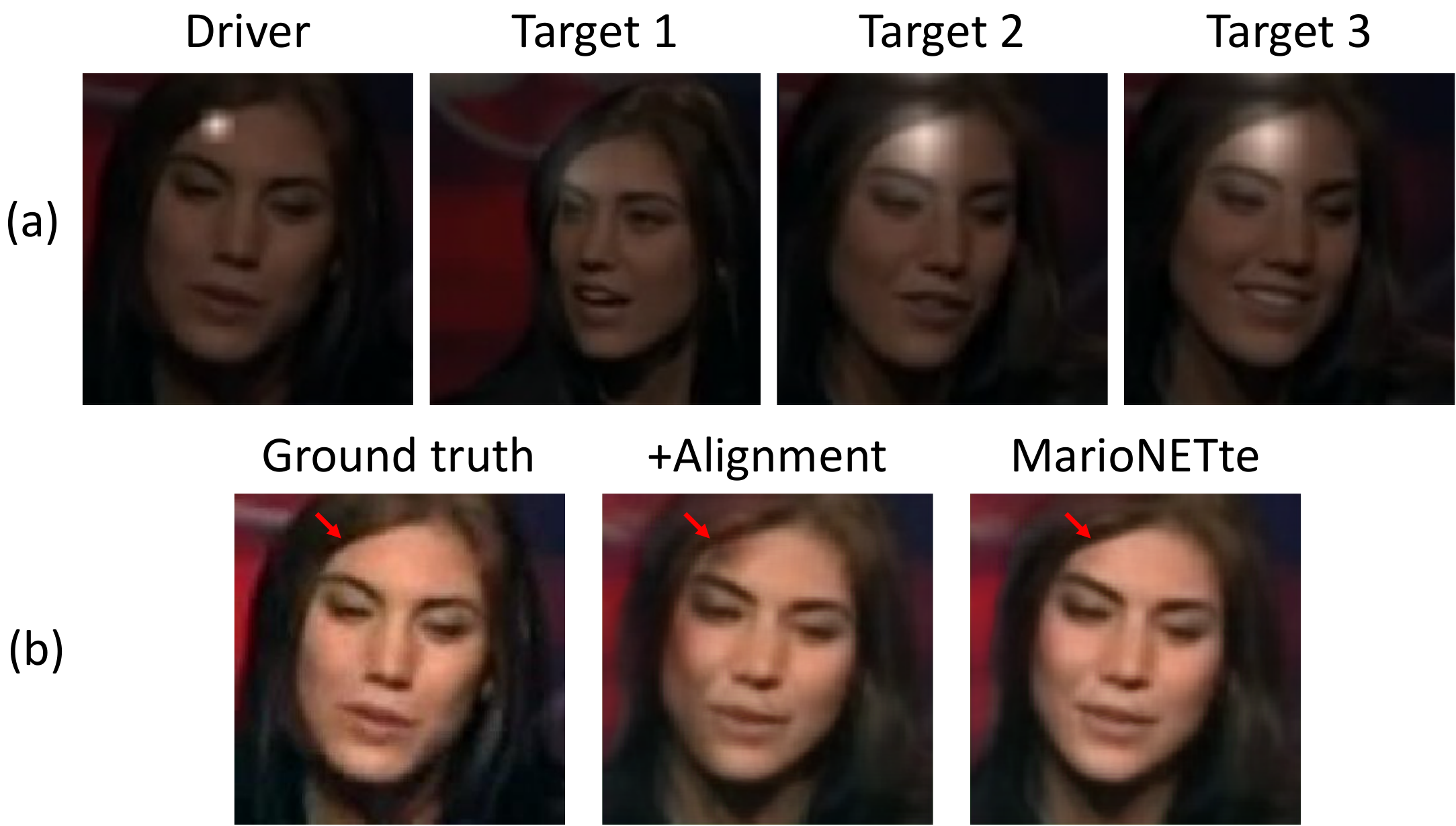}
    \caption{(a) Driver and target images overlapped with attention map. Brightness signifies the intensity of the attention. 
    (b) Failure case of \textit{+Alignment} and improved result generated by \textit{MarioNETte}.}
    \label{fig:ablation_qualitive}
 \end{figure}

We performed ablation test to investigate the effectiveness of the proposed components.
While keeping all other things the same, we compare the following configurations reenacting different identities: (1) \textbf{MarioNETte} is the proposed method where both image attention block and target feature alignment are applied. (2) \textbf{AdaIN} corresponds to the same model as MarioNETte, where the image attention block is replaced with AdaIN residual block while the target feature alignment is omitted. (3) \textbf{+Attention} is a MarioNETte where only the image attention block is applied. (4) \textbf{+Alignment} only employs the target feature alignment.

Table~\ref{tab:ablation_result} shows result of ablation test.
For identity preservation (i.e., CSIM), \textit{AdaIN} has a hard time combining style features depending solely on AdaIN residual blocks.
\textit{+Attention} alleviates the problem immensely in both one-shot and few-shot settings by attending to proper coordinates.
While \textit{+Alignment} exhibits a higher CSIM compared to \textit{+Attention}, it struggles in generating plausible images for unseen poses and expressions leading to worse PRMSE and AUCON.
Taking advantage of both attention and target feature alignment, \textit{MarioNETte} outperforms \textit{+Alignment} in every metric under consideration.

Entirely relying on target feature alignment for reenactment, \textit{+Alignment} is vulnerable to failures due to large differences in pose between target and driver that \textit{MarioNETte} can overcome.
Given a single driver image along with three target images (Figure~\ref{fig:ablation_qualitive}a), \textit{+Alignment} has defects on the forehead (denoted by arrows in Figure~\ref{fig:ablation_qualitive}b).
This is due to (1) warping low-level features from a large-pose input and (2) aggregating features from multiple targets with diverse poses.
\textit{MarioNETte}, on the other hand, gracefully handles the situation by attending to proper image among several target images as well as adequate spatial coordinates in the target image.
The attention map, highlighting the area where the image attention block is focusing on, is illustrated with white in Figure~\ref{fig:ablation_qualitive}a.
Note that \textit{MarioNETte} attends to the forehead and  adequate target images (Target 2 and 3 in Figure~\ref{fig:ablation_qualitive}a) which has similar pose with driver.

\section{Related Works}

The classical approach to face reenactment commonly involves the use of explicit 3D modeling of human faces~\cite{blanz-siggraph-1999-3dmm} where the 3DMM parameters of the driver and the target are computed from a single image, and blended eventually~\cite{thies-siggraphasia-2015-realtimeexpression,thies-cvpr-2016-face2face}.
Image warping is another popular approach where the target image is modified using the estimated flow obtained form 3D models~\cite{cao-ieee-2013-facewarehouse} or sparse landmarks~\cite{averbuch-2017-siggraph-portraitstolife}.
Face reenactment studies have embraced the recent success of neural networks exploring different image-to-image translation architectures~\cite{isola-cvpr-2017-pix2pix} such as the works of \citet{xu-arxiv-2017-facetransfer} and that of \citet{wu-eccv-2018-reenactgan}, which combined the cycle consistency loss~\cite{zhu-cvpr-2017-cyclegan}.
A hybrid of two approaches has been studied as well.
\citet{kim-siggraph-2018-deepvideoportraits} trained an image translation network which maps reenacted render of a 3D face model into a photo-realistic output.

Architectures, capable of blending the style information of the target with the spatial information of the driver, have been proposed recently.
AdaIN~\cite{huang-cvpr-2017-adain,huang-eccv-2018-munit,liu-arxiv-2019-funit} layer, attention mechanism~\cite{zhu-2019-cvpr-progressiveattention,lathuiliere-arxiv-2019-attentionfusion,park-arxiv-2019-styleattention}, deformation operation~\cite{siarohin-cvpr-2018-deformablegan,dong-2018-nips-softgatedgan}, and GAN-based method~\cite{bao2018towards} have all seen a wide adoption.
Similar idea has been applied to few-shot face reenactment settings such as the use of image-level~\cite{wiles-eccv-2018-x2face} and feature-level~\cite{siarohin-cvpr-2019-monkeynet} warping, and AdaIN layer in conjuction with a meta-learning~\cite{zakharov-arxiv-2019-samsung}.
The identity mismatch problem has been studied through methods such as CycleGAN-based landmark transformers~\cite{wu-eccv-2018-reenactgan} and landmark swappers~\cite{zhang-arxiv-2019-faceswapnet}.
While effective, these methods either require an independent model per person or a dataset with image pairs that may be hard to acquire.

\section{Conclusions}

In this paper, we have proposed a framework for few-shot face reenactment.
Our proposed image attention block and target feature alignment, together with the landmark transformer, allow us to handle the identity mismatch caused by using the landmarks of a different person.
Proposed method do not need additional fine-tuning phase for identity adaptation, which significantly increases the usefulness of the model when deployed in-the-wild.
Our experiments including human evaluation suggest the excellence of the proposed method.

One exciting avenue for future work is to improve the landmark transformer to better handle the landmark disentanglement to make the reenactment even more convincing.

\bibliographystyle{aaai}
\fontsize{9.0pt}{10.0pt} \selectfont
\bibliography{6371.MarioNETte.bib}

\clearpage

\appendixpageoff
\appendixtitleoff
\renewcommand{\appendixtocname}{Supplementary Material}
\begin{appendices}
    \begin{center}
    \textbf{\large Supplemental Materials}
    \end{center}

\section{MarioNETte Architecture Details}

\subsection{Architecture design}

Given a driver image $\mathbf{x}$ and $K$ target images $\{\mathbf{y}^{i}\}_{i=1 \ldots K}$, the proposed few-shot face reenactment framework which we call \textit{MarioNETte} first generates 2D landmark images (i.e. $\mathbf{r}_x$ and $\{\mathbf{r}_y^{i}\}_{i=1 \ldots K}$).
We utilize a \textit{3D landmark detector} $\mathcal{K}:\mathbb{R}^{h \times w \times 3} \xrightarrow{} \mathbb{R}^{68 \times 3}$~\cite{bulat-iccv-2017-facealignment} to extract facial keypoints which includes information about pose and expression denoted as $\mathbf{l}_x = \mathcal{K}(\mathbf{x})$ and $\mathbf{l}_{y}^{i} = \mathcal{K}(\mathbf{y}^{i})$, respectively.
We further rasterize 3D landmarks to an image by rasterizer $\mathcal{R}$, resulting in $\mathbf{r}_x = \mathcal{R}(\mathbf{l}_x), \mathbf{r}_{y}^{i} = \mathcal{R}(\mathbf{l}_{y}^{i})$.

\begin{figure}[h]
 \centering
    \includegraphics[width=0.7\columnwidth]{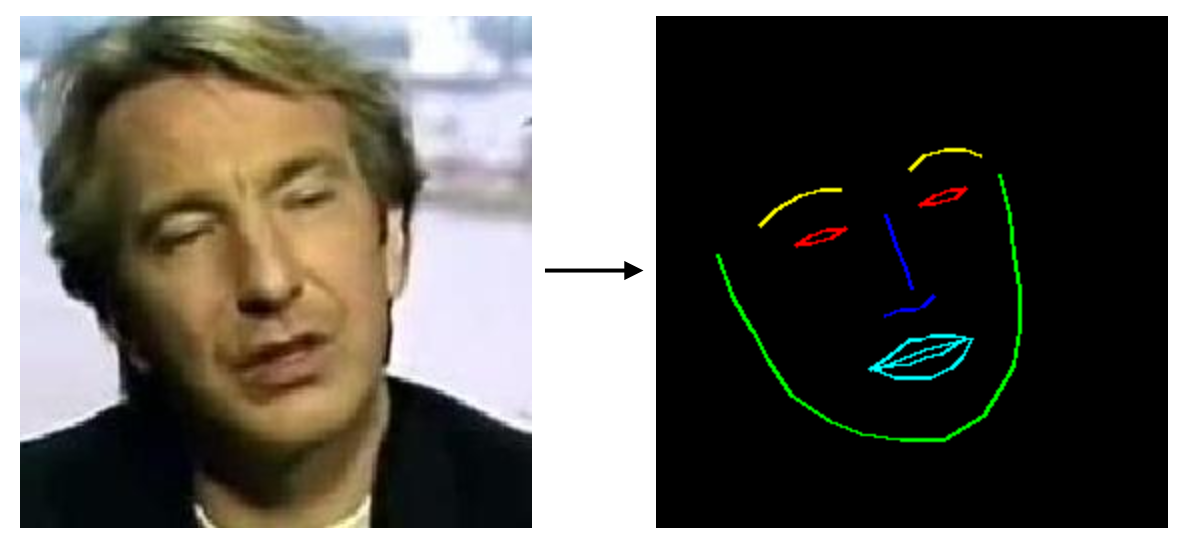}
    \caption{Example of the rasterized facial landmarks.}
    \label{fig:example_landmark_rasterize}
 \end{figure}

We utilize simple rasterizer that orthogonally projects 3D landmark points, e.g., $(x, y, z)$, into 2D $XY$-plane, e.g., $(x, y)$, and we group the projected landmarks into 8 categories: left eye, right eye, contour, nose, left eyebrow, right eyebrow, inner mouth, and outer mouth.
For each group, lines are drawn between predefined order of points with predefined colors (e.g., red, red, green, blue, yellow, yellow, cyan, and cyan respectively), resulting in a rasterized image as shown in Figure~\ref{fig:example_landmark_rasterize}.

MarioNETte consists of \textbf{conditional image generator} $G(\mathbf{r}_x;\{\mathbf{y}^{i}\}_{i=1 \ldots K}, \{\mathbf{r}_{y}^{i}\}_{i=1 \ldots K})$ and \textbf{projection discriminator} $D(\mathbf{\hat{x}}, \mathbf{\hat{r}}, c)$. 
The discriminator $D$ determines whether the given image $\mathbf{\hat{x}}$ is a real image from the data distribution taking into account the conditional input of the rasterized landmarks $\mathbf{\hat{r}}$ and identity $c$.

The \textbf{generator} $G(\mathbf{r}_x;\{\mathbf{y}^{i}\}_{i=1 \ldots K}, \{\mathbf{r}_{y}^{i}\}_{i=1 \ldots K})$ is further broken down into four components: namely, \textit{target encoder}, \textit{drvier encoder}, \textit{blender}, and \textit{decoder}.
\textit{Target encoder} $E_y(\mathbf{y}, \mathbf{r}_y)$ takes target image and generates encoded target feature map $\mathbf{z}_y$ together with the warped target feature map $\hat{\mathbf{S}}$.
\textit{Driver encoder} $E_x(\mathbf{r}_x)$ receives a driver image and creates a driver feature map $\mathbf{z}_x$.
\textit{Blender} $B(\mathbf{z}_x, \{\mathbf{z}_{y}^{i}\}_{i=1 \ldots K})$ combines encoded feature maps to produce a mixed feature map $\mathbf{z}_{xy}$.
\textit{Decoder} $Q(\mathbf{z}_{xy}, \{\hat{\mathbf{S}}^{i}\}_{i=1 \ldots K})$ generates the reenacted image.
Input image $\mathbf{y}$ and the landmark image $\mathbf{r}_y$ are concatenated channel-wise and fed into the target encoder.

The \textbf{target encoder} $E_y(\mathbf{y}, \mathbf{r}_y)$ adopts a U-Net~\cite{ronneberger-miccai-2015-unet} style architecture including five downsampling blocks and four upsampling blocks with skip connections.
Among five feature maps $\{\mathbf{s}_{j}\}_{j=1 \ldots 5}$ generated by the downsampling blocks, the most downsampled feature map, $\mathbf{s}_5$, is used as the encoded target feature map $\mathbf{z}_y$, while the others, $\{\mathbf{s}_{j}\}_{j=1 \ldots 4}$, are transformed into normalized feature maps.
A normalization flow map $\mathbf{f}_{y} \in \mathbb{R}^{(h/2) \times (w/2) \times 2}$ transforms each feature map into normalized feature map, $\hat{\mathbf{S}} = \{\hat{\mathbf{s}}_{j}\}_{j=1 \ldots 4}$, through warping function $\mathcal{T}$ as follows:
\begin{equation}
    \hat{\mathbf{S}} = \{\mathcal{T}(\mathbf{s}_{1}; \mathbf{f}_{y}), \ldots ,\mathcal{T}(\mathbf{s}_{4}; \mathbf{f}_{y})\}.
\end{equation}
\noindent Flow map $\mathbf{f}_y$ is generated at the end of upsampling blocks followed by an additional convolution layer and a hyperbolic tangent activation layer, thereby producing a 2-channel feature map, where each channel denotes a flow for the horizontal and vertical direction, respectively.

We adopt bilinear sampler based warping function which is widely used along with neural networks due to its differentiability~\cite{jaderberg-nips-2015-stn,balakrishnan-cvpr-2018-synthesizing,siarohin-cvpr-2019-monkeynet}.
Since each $\mathbf{s}_{j}$ has a different width and height, average pooling is applied to downsample $\mathbf{f}_{y}$ to match the size of $\mathbf{f}_{y}$ to that of $\mathbf{s}_{j}$.

The \textbf{driver encoder} $E_x(\mathbf{r}_x)$, which consists of four residual downsampling blocks, takes driver landmark image $\mathbf{r}_x$ and generates driver feature map $\mathbf{z}_x$.

The \textbf{blender} $B(\mathbf{z}_x, \{\mathbf{z}_{y}^{i}\}_{i=1 \ldots K})$ produces mixed feature map $\mathbf{z}_{xy}$ by blending the positional information of $\mathbf{z}_x$ with the target style feature maps $\mathbf{z}_y$.
We stacked three \textit{image attention blocks} to build our blender.

The \textbf{decoder} $Q(\mathbf{z}_{xy}, \{\hat{\mathbf{S}}^{i}\}_{i=1 \ldots K})$ consists of four \textit{warp-alignment blocks} followed by residual upsampling blocks.
Note that the last upsampling block is followed by an additional convolution layer and a hyperbolic tangent activation function.

The \textbf{discriminator} $D(\mathbf{\hat{x}}, \mathbf{\hat{r}}, c)$ consists of five residual downsampling blocks without self-attention layers.
We adopt a projection discriminator with a slight modification of removing the global sum-pooling layer from the original structure.
By removing the global sum-pooling layer, discriminator generates scores on multiple patches like PatchGAN discriminator~\cite{isola-cvpr-2017-pix2pix}.

We adopt the residual upsampling and downsampling block proposed by \citet{brock-iclr-2019-biggan} to build our networks.
All batch normalization layers are substituted with instance normalization except for the target encoder and the discriminator, where the normalization layer is absent.
We utilized ReLU as an activation function.
The number of channels is doubled (or halved) when the output is downsampled (or upsampled).
The minimum number of channels is set to 64 and the maximum number of channels is set to 512 for every layer.
Note that the input image, which is used as an input for the target encoder, driver encoder, and discriminator, is first projected through a convolutional layer to match the channel size of 64.

\subsection{Positional encoding}

We utilize a sinusoidal positional encoding introduced by~\citet{vaswani-nips-2017-transformer} with a slight modification. 
First, we divide the number of channels of the positional encoding in half.
Then, we utilize half of them to encode the horizontal coordinate and the rest of them to encode the vertical coordinate. 
To encode the relative position, we normalize the absolute coordinate by the width and the height of the feature map.
Thus, given a feature map of $\mathbf{z} \in \mathbb{R}^{h_z \times w_z \times c_z}$, the corresponding positional encoding $\mathbf{P} \in \mathbb{R}^{h_z \times w_z \times c_z}$ is computed as follows:

\begin{equation}
    \begin{aligned}
    \mathbf{P}_{i, j, 4k} = & \sin\left(\frac{256i}{h_z \cdot 10000^{2k / c_z}}\right) \\
    \mathbf{P}_{i, j, 4k+1} = & \cos\left(\frac{256i}{h_z \cdot 10000^{2k / c_z}}\right) \\
    \mathbf{P}_{i, j, 4k+2} = & \sin\left(\frac{256j}{w_z \cdot 10000^{2k / c_z}}\right) \\
    \mathbf{P}_{i, j, 4k+3} = & \cos\left(\frac{256j}{w_z \cdot 10000^{2k / c_z}}\right).
    \end{aligned}
\end{equation}

\subsection{Loss functions}

Our model is trained in an adversarial manner using a projection discriminator $D$~\cite{miyato-arxiv-2018-projdisc}.
The discriminator aims to distinguish between the real image of the identity $c$ and a synthesized image of $c$ generated by $G$.
Since the paired target and the driver images from different identities cannot be acquired without explicit annotation, we trained our model using the target and the driver image extracted from the same video.
Thus, identities of $\mathbf{x}$ and $\mathbf{y}^i$ are always the same, e.g., $c$, for every target and driver image pair, i.e., $(\mathbf{x}, \{\mathbf{y}^i\}_{i=1 \ldots K})$, during the training.

We use hinge GAN loss~\cite{lim-2017-arxiv-geometricgan} to optimize discriminator $D$ as follows:
\begin{equation}
\begin{aligned}
    \mathbf{\hat{x}} = & \quad G(\mathbf{r}_x;\{\mathbf{y}^{i}\}, \{\mathbf{r}_{y}^{i}\}) \\
    \mathcal{L}_D = & \quad \textrm{max}(0, 1 - D(\mathbf{x}, \mathbf{r}_x, c)) \quad + \\
    & \quad \textrm{max}(0, 1 + D(\mathbf{\hat{x}}, \mathbf{r}_x, c)).
\end{aligned}
\end{equation}

The loss function of the generator consists of four components including the GAN loss $\mathcal{L}_{GAN}$, the perceptual losses ($\mathcal{L}_P$ and $\mathcal{L}_{PF}$), and the feature matching loss $\mathcal{L}_{FM}$.
The GAN loss $\mathcal{L}_{GAN}$ is a generator part of the hinge GAN loss and defined as follows:
\begin{equation}
\begin{aligned}
    \mathcal{L}_{GAN} = -D(\mathbf{\hat{x}}, \mathbf{r}_x, c).
\end{aligned}
\end{equation}

\noindent The perceptual loss~\cite{johnson-eccv-2016-perceptual-loss} is calculated by averaging $L_{1}$-distances between the intermediate features of the pre-trained network using ground truth image $\mathbf{x}$ and the generated image $\mathbf{\hat{x}}$.
We use two different networks for perceptual losses where $\mathcal{L}_P$ and $\mathcal{L}_{PF}$ are extracted from VGG19 and VGG-VD-16 each trained for ImageNet classification task~\cite{simonyan-2014-arxiv-vgg} and a face recognition task~\cite{parkhi-bmvc-2015-vggface}, respectively.
We use features from the following layers to compute the perceptual losses: \texttt{relu1\_1}, \texttt{relu2\_1}, \texttt{relu3\_1}, \texttt{relu4\_1}, and \texttt{relu5\_1}.
Feature matching loss $\mathcal{L}_{FM}$ is the sum of $L_{1}$-distances between the intermediate features of the discriminator $D$ when processing the ground truth image $\mathbf{x}$ and the generated image $\mathbf{\hat{x}}$ which helps with the stabilization of the adversarial training.
It helps to stabilize the adversarial training.
The overall generator loss is the weighted sum of the four losses:
\begin{equation}
    \mathcal{L}_G = \mathcal{L}_{GAN} + \lambda_P \mathcal{L}_P + \lambda_{PF} \mathcal{L}_{PF} + \lambda_{FM} \mathcal{L}_{FM}.
\end{equation}

\subsection{Training details}

To stabilize the adversarial training, we apply spectral normalization~\cite{miyato-iclr-2018-spectralnorm} for every layer of the discriminator and the generator.
In addition, we use the convex hull of the facial landmarks as a facial region mask and give three-fold weights to the corresponding masked position while computing the perceptual loss.
We use Adam optimizer to train our model where the learning rate of $2\times10^{-4}$ is used for the discriminator and $5\times10^{-5}$ is used for the generator and the style encoder. 
Unlike the setting of \citet{brock-iclr-2019-biggan}, we only update the discriminator once per every generator updates.
We set $\lambda_P$ to 10, $\lambda_{PF}$ to 0.01, $\lambda_{FM}$ to 10, and the number of target images $K$ to 4 during the training.

\section{Landmark Transformer Details}

\subsection{Landmark decomposition}

Formally, landmark decomposition is calculated as:
\begin{equation}
\begin{aligned}
    \mathbf{\bar{l}}_m &= \frac{1}{CT} \sum_{c} \sum_{t} \mathbf{\bar{l}}(c,t), \\
    \mathbf{\bar{l}}_{id}(c) &= \frac{1}{T_c} \sum_{t} \mathbf{\bar{l}}(c,t) - \mathbf{\bar{l}}_m, \\
    \mathbf{\bar{l}}_{exp}(c,t) &= \mathbf{\bar{l}}(c,t) - \mathbf{\bar{l}}_m - \mathbf{\bar{l}}_{id}(c) \\
    \quad &= \mathbf{\bar{l}}(c,t) - \frac{1}{T_c} \sum_{t} \mathbf{\bar{l}}(c,t), \\
\end{aligned}
\label{eq:suppl_landmark_decomposition}
\end{equation}
where $C$ is the number of videos, $T_c$ is the number of frames of $c$-th video, and $T = \sum T_{c}$.
We can easily compute the components shown in Equation~\ref{eq:suppl_landmark_decomposition} from the training dataset.

However, when an image of unseen identity $c^\prime$ is given, the decomposition of the identity and the expression shown in Equation~\ref{eq:suppl_landmark_decomposition} is not possible since $\mathbf{\bar{l}}_{exp}(c^\prime,t)$ will be zero for a single image.
Even when a few frames of an unseen identity $c^\prime$ is given, $\mathbf{\bar{l}}_{exp}(c^\prime,t)$ will be zero (or near zero) if the expressions in the given frames are not diverse enough.
Thus, to perform the decomposition shown in Equation~\ref{eq:suppl_landmark_decomposition} even under the one-shot or few-shot settings, we introduce \textit{landmark disentangler}.

\subsection{Landmark disentanglement}

To compute the expression basis $\mathbf{b}_{exp}$, using the expression geometry obtained from the VoxCeleb1 training data, we divide a landmark into different groups (e.g., left eye, right eye, eyebrows, mouth, and any other) and perform PCA on each group.
We utilize PCA dimensions of 8, 8, 8, 16 and 8, for each group, resulting in a total number of expression bases, $n_{exp}$, of 48.

We train landmark disentangler on the VoxCeleb1 training set, separately.
Before training landmark disentangler, we normalized each expression parameter $\alpha_i$ to follow a standard normal distribution $\mathcal{N}(0, 1^2)$ for the ease of regression training.
We employ ResNet50, which is pre-trained on ImageNet~\cite{he-2016-cvpr-deep}, and extract features from the first layer to the last layer right before the global average pooling layer.
Extracted image features are concatenated with the normalized landmark $\mathbf{\bar{l}}$ subtracted by the mean landmark $\mathbf{\bar{l}}_m$, and fed into a 2-layer MLP followed by a ReLU activation.
The whole network is optimized by minimizing the MSE loss between the predicted expression parameters and the target expression parameters, using Adam optimizer with a learning rate of $3\times10^{-4}$.
We use gradient clipping with the maximum gradient norm of 1 during the training.
We set the expression intensity parameter $\lambda_{exp}$ to 1.5.

\section{Additional Ablation Tests}

\subsection {Quantitative results}

In Table 1 and Table 2 of the main paper, \textit{MarioNETte} shows better PRMSE and AUCON under the self-reenactment setting on VoxCeleb1 compared to \textit{NeuralHead-FF}, which, however, is reversed under the reenactment of a different identity on CelebV.
We provide an explanation of this phenomenon through an ablation study.

Table~\ref{tab:ablation_self_result} illustrates the evaluation results of ablation models under self-reenactment settings on VoxCeleb1.
Unlike the evaluation results of reenacting a different identity on CelebV (Table 4 of the main paper), \textit{+Alignment} and \textit{MarioNETte} show better PRMSE and AUCON compared to the \textit{AdaIN}.
The phenomenon may be attributed to the characteristics of the training dataset as well as the different inductive biases of different models.
VoxCeleb1 consists of short video clips (usually 5-10s long), leading to similar poses and expressions between drivers and targets.
Unlike the AdaIN-based model which is unaware of spatial information, the proposed image attention block and the target feature alignment encode spatial information from the target image.
We suspect that this may lead to possible overfitting of the proposed model to the \textit{same identity pair with a similar pose and expression} setting.

\begin{table}[t]
    \centering
    \begin{tabular}{cccc}
        \toprule
Model (\# target) & CSIM$\uparrow$ & PRMSE$\downarrow$ & AUCON$\uparrow$ \\
\midrule
AdaIN (1) & 0.183 & 3.719 & 0.781 \\
+Attention (1) & 0.611 & 3.257 & \underline{0.825} \\
+Alignment (1) & \bf{0.756} & \bf{3.069} & \bf{0.827} \\
\midrule
MarioNETte (1) & \underline{0.755} & \underline{3.125} & \underline{0.825} \\
\midrule
\midrule
AdaIN (8) & 0.188 & 3.649 & 0.787 \\
+Attention (8) & 0.717 & 2.909 & 0.843 \\
+Alignment (8) & \underline{0.826} & \bf{2.563} & \underline{0.845} \\
\midrule
MarioNETte (8) & \bf{0.828} & \underline{2.571} & \bf{0.850} \\
\bottomrule
    \end{tabular}
    \caption{Comparison of ablation models for self-reenactment setting on VoxCeleb1 dataset.}
    \label{tab:ablation_self_result}
\end{table}

\subsection {Qualitative results}

\begin{figure*}[t]
    \includegraphics[width=0.99\linewidth]{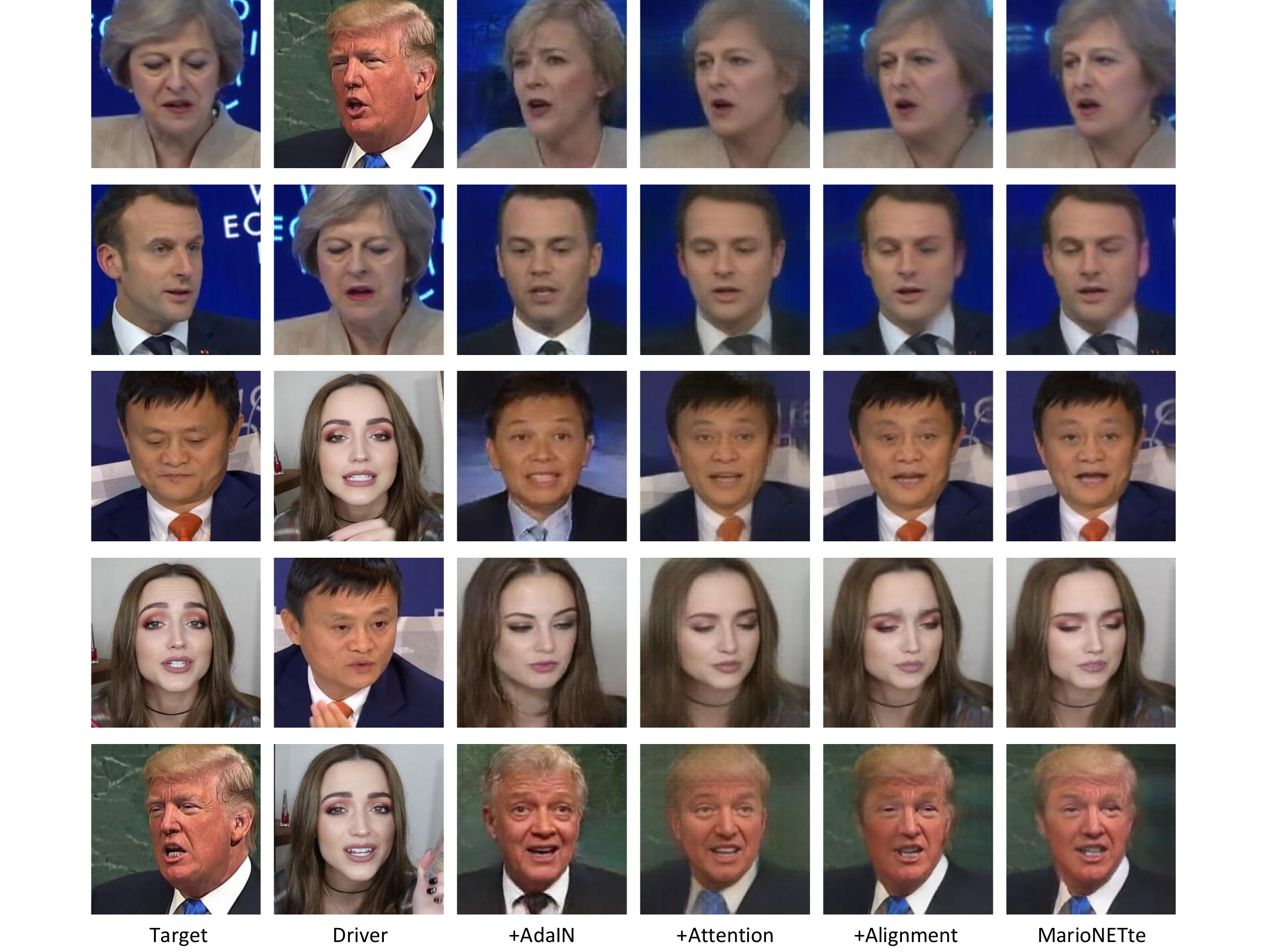}
    \caption{Qualitative results of ablation models of one-shot reenactment under different identity setting on CelebV.}
    \label{fig:celebv_diff_id_reenact_ablation_1ref}
\end{figure*}

\begin{figure*}[t]
    \includegraphics[width=0.99\linewidth]{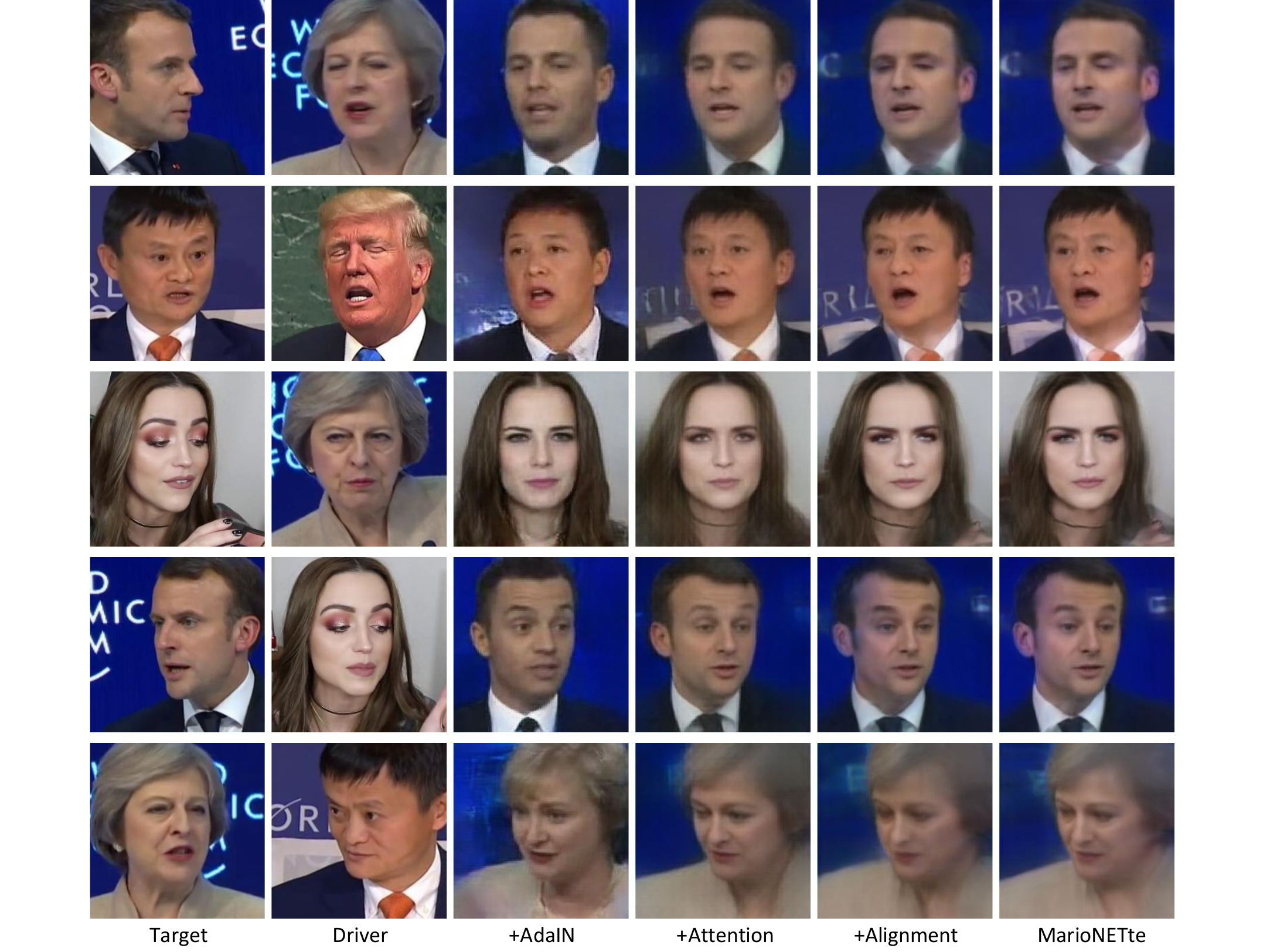}
    \caption{Qualitative results of ablation models of few-shot reenactment under different identity setting on CelebV.}
    \label{fig:celebv_diff_id_reenact_ablation_8ref}
\end{figure*}

Figure~\ref{fig:celebv_diff_id_reenact_ablation_1ref} and Figure~\ref{fig:celebv_diff_id_reenact_ablation_8ref} illustrate the results of ablation models reenacting a different identity on CelebV under the one-shot and few-shot settings, respectively.
While \textit{AdaIN} fails to generate an image that resembles the target identity, \textit{+Attention} successfully maintains the key characteristics of the target.
The target feature alignment module adds fine-grained details to the generated image.
However, \textit{MarioNETte} tends to generate more natural images in a few-shot setting, while \textit{+Alignment} struggles to deal with multiple target images with diverse poses and expressions.

\section{Inference Time}

In this section, we report the inference time of our model.
We measured the latency of the proposed method while generating $256\times256$ images with different number of target images, K $\in\{1, 8\}$.
We ran each setting for 300 times and report the average speed.
We utilized \textit{Nvidia Titan Xp} and \textit{Pytorch 1.0.1.post2}.
As mentioned in the main paper, we used the open-sourced implementation of \citet{bulat-iccv-2017-facealignment} to extract 3D facial landmarks.

\begin{table*}
    \centering
    \begin{tabular}{ccc}
        \toprule
Description & Symbol & Inference time (ms) \\
\midrule
3D Landmark Detector & $T_{P}$ & 101 \\
Target Encoder & $T_{E, K}$ & 44 (K=1), 111 (K=8) \\
Target Landmark Transformer & $T_{TLT, K}$ & 22 (K=1), 19 (K=8) \\ 
Generator & $T_{G, K}$ & 35 (K=1), 36 (K=8) \\ 
Driver Landmark Transformer & $T_{DLT}$ & 26 \\
        \bottomrule
    \end{tabular}
    \caption{Inference speed of each component of our model.}
    \label{tab:inference_speed_of_components}
\end{table*} 

\begin{table*}
    \centering
    \begin{tabular}{ccc}
        \toprule
Model & Target encoding & Driver generation \\
\midrule
MarioNETte+LT & $K \cdot T_{P} + T_{TLT, K} + T_{E, K}$ & $T_{P} + T_{DLT} + T_{G, K}$ \\
MarioNETte & $K \cdot T_{P} + T_{E, K}$ & $T_{P} + T_{G, K}$ \\
        \bottomrule
    \end{tabular}
    \caption{Inference speed of the full model for generating single image with $K$ target images.}
    \label{tab:inference_speed_of_full_model}
\end{table*}

Table~\ref{tab:inference_speed_of_components} displays the inference time breakdown of our models.
Total inference time of the proposed models, \textit{MarioNETte+LT} and \textit{MarioNETte}, can be derived as shown in Table~\ref{tab:inference_speed_of_full_model}.
While generating reenactment videos, $z_y$ and $\hat{\mathbf{S}}$, utilized to compute the target encoding, is generated only once at the beginning. 
Thus, we divide our inference pipeline into \textit{Target encoding} part and the \textit{Driver generation} part.

Since we perform a batched inference for multiple target images, the inference time of the proposed components (e.g., the target encoder and the target landmark transformer) scale sublinearly to the number of target images $K$.
On the other hand, the open-source 3D landmark detector processes images in a sequential manner, and thus, its processing time scales linearly.

\section{Additional Examples of Generated Images}

We provide additional qualitative results of the baseline methods and the proposed models on VoxCeleb1 and CelebV datasets.
We report the qualitative results for both one-shot and few-shot (8 target images) settings, except \textit{Monkey-Net} which is designed for using only a single image.
In the case of the few-shot reenactment, we display only one target image, due to the limited space.

\begin{figure*}[t]
    \includegraphics[width=0.99\linewidth]{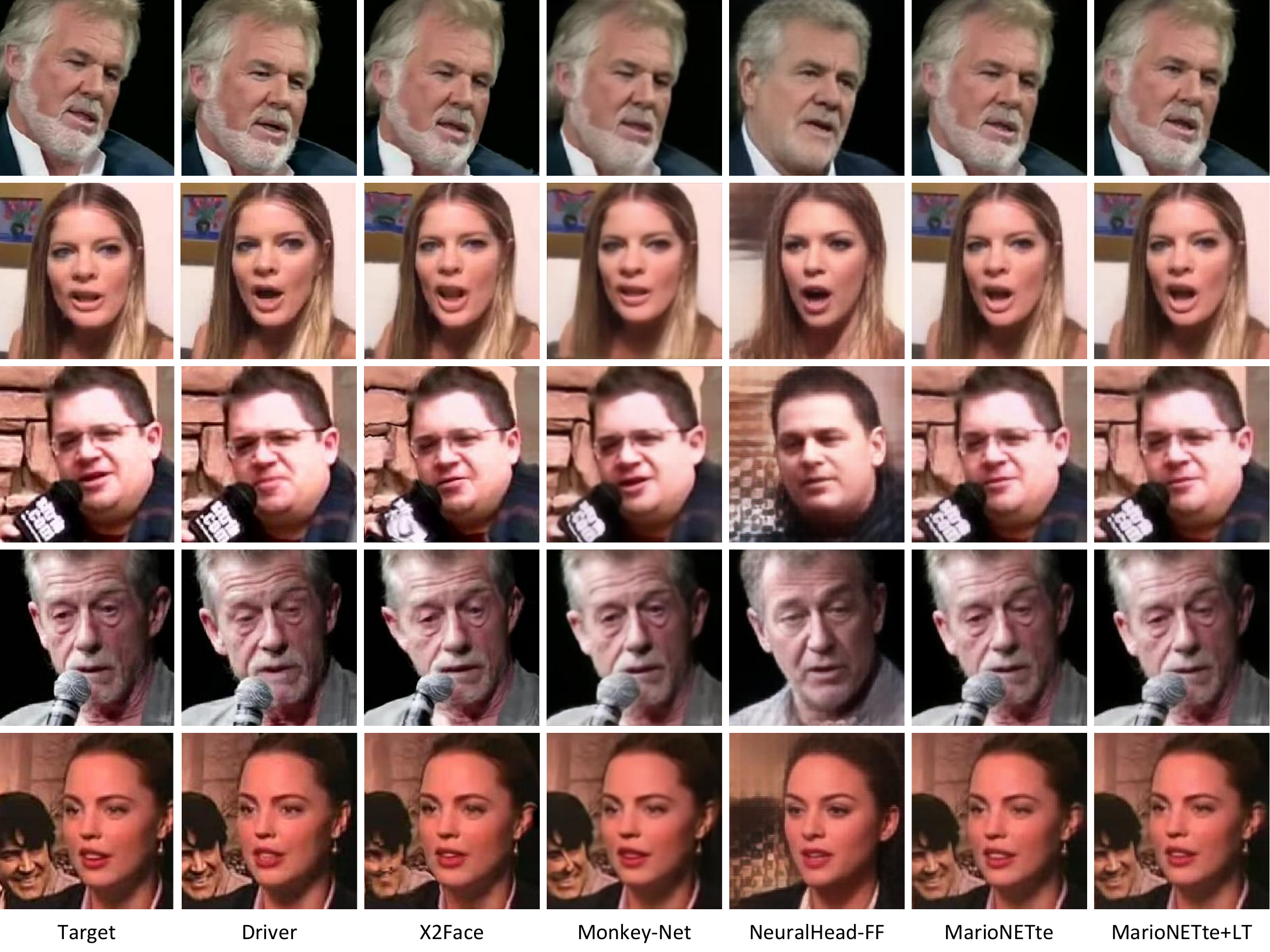}
    \caption{Qualitative results of one-shot self-reenactment setting on VoxCeleb1.}
    \label{fig:vox_same_id_realism_1ref}
\end{figure*}

\begin{figure*}[t]
    \includegraphics[width=0.99\linewidth]{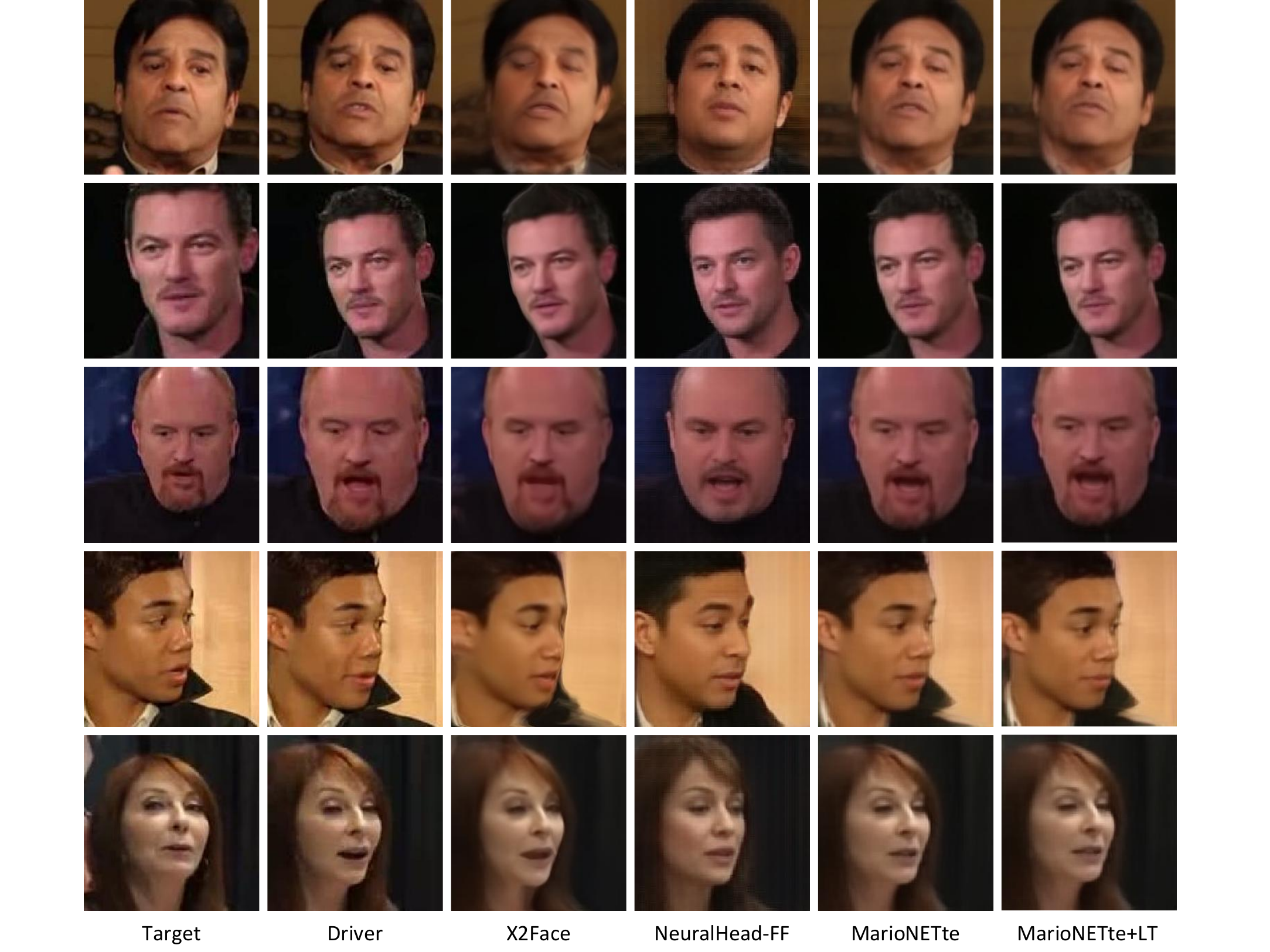}
    \caption{Qualitative results of few-shot self-reenactment setting on VoxCeleb1.}
    \label{fig:vox_same_id_reenact_8ref}
\end{figure*}

\begin{figure*}[t]
    \includegraphics[width=0.99\linewidth]{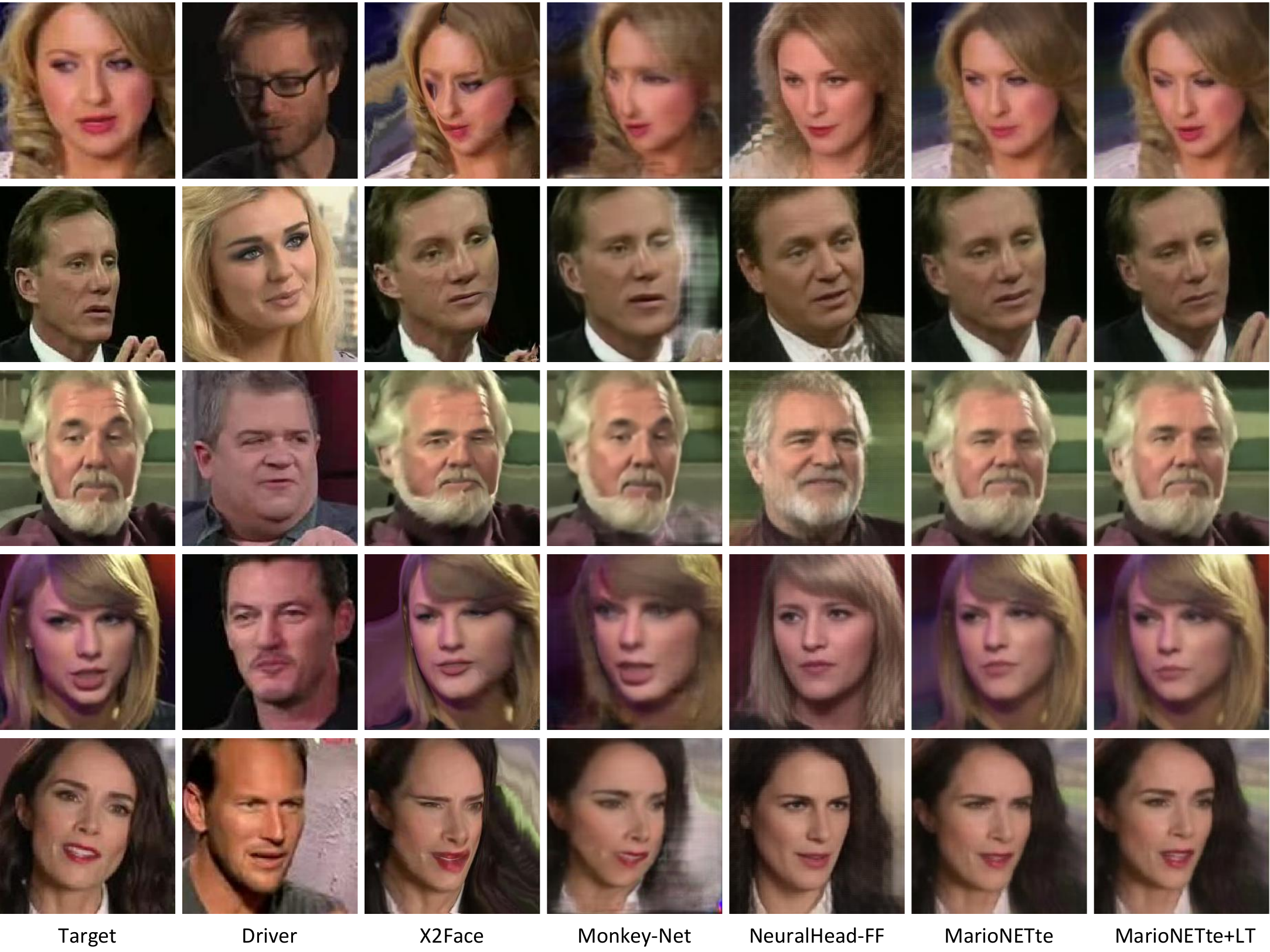}
    \caption{Qualitative results of one-shot reenactment under different identity setting on VoxCeleb1.}
    \label{fig:vox_diff_id_realism_1ref}
\end{figure*}

\begin{figure*}[t]
    \includegraphics[width=0.99\linewidth]{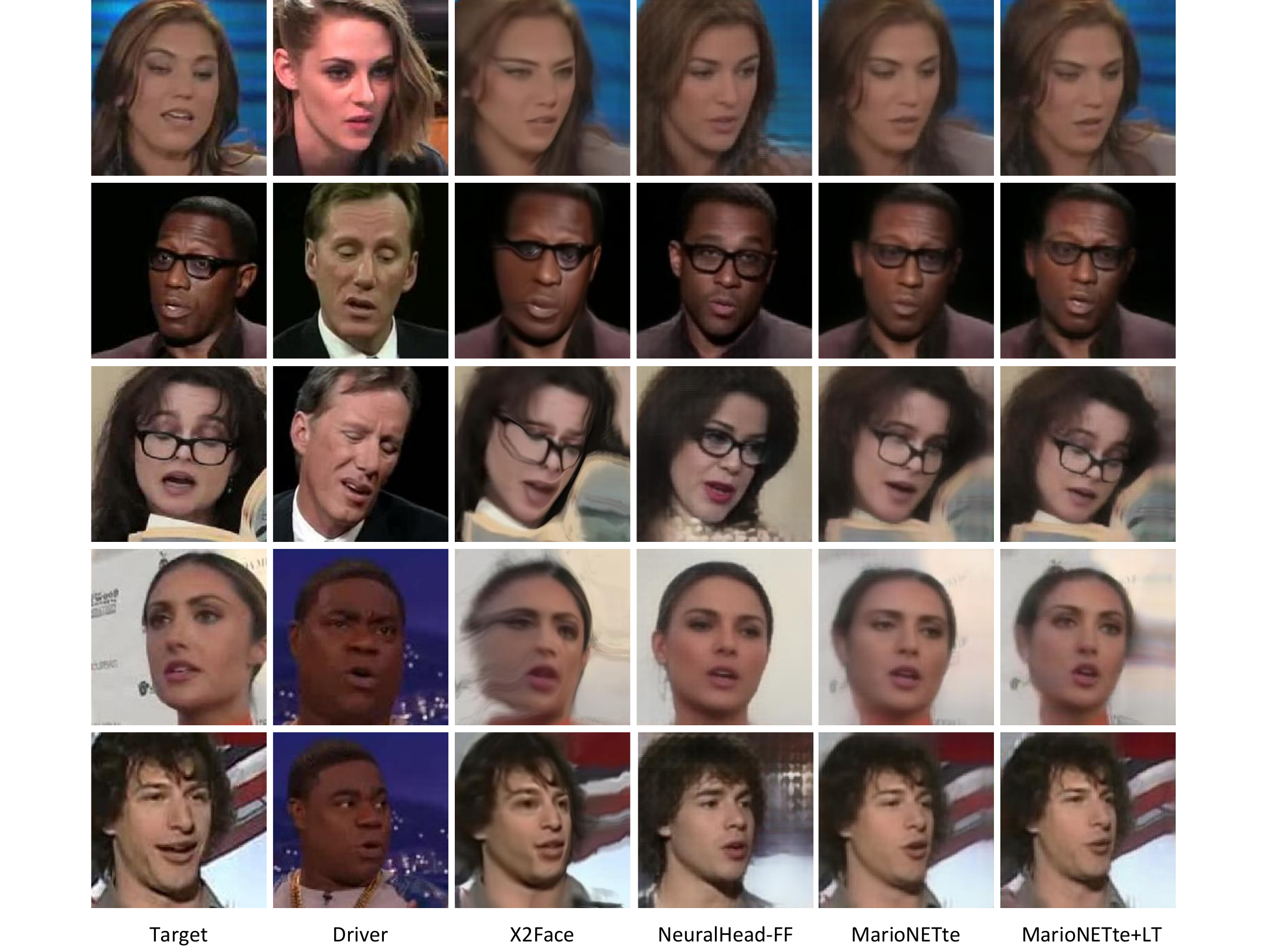}
    \caption{Qualitative results of few-shot reenactment under different identity setting on VoxCeleb1.}
    \label{fig:vox_diff_id_realism_8ref}
\end{figure*}

Figure~\ref{fig:vox_same_id_realism_1ref} and Figure~\ref{fig:vox_same_id_reenact_8ref} compare different methods for the self-reenactment on VoxCeleb1 in one-shot and few-shot settings, respectively.
Examples of one-shot and few-shot reenactments on VoxCeleb1 where driver's and target's identity do not match is shown in Figures~\ref{fig:vox_diff_id_realism_1ref} and Figure~\ref{fig:vox_diff_id_realism_8ref}, respectively.

\begin{figure*}[t]
    \includegraphics[width=0.99\linewidth]{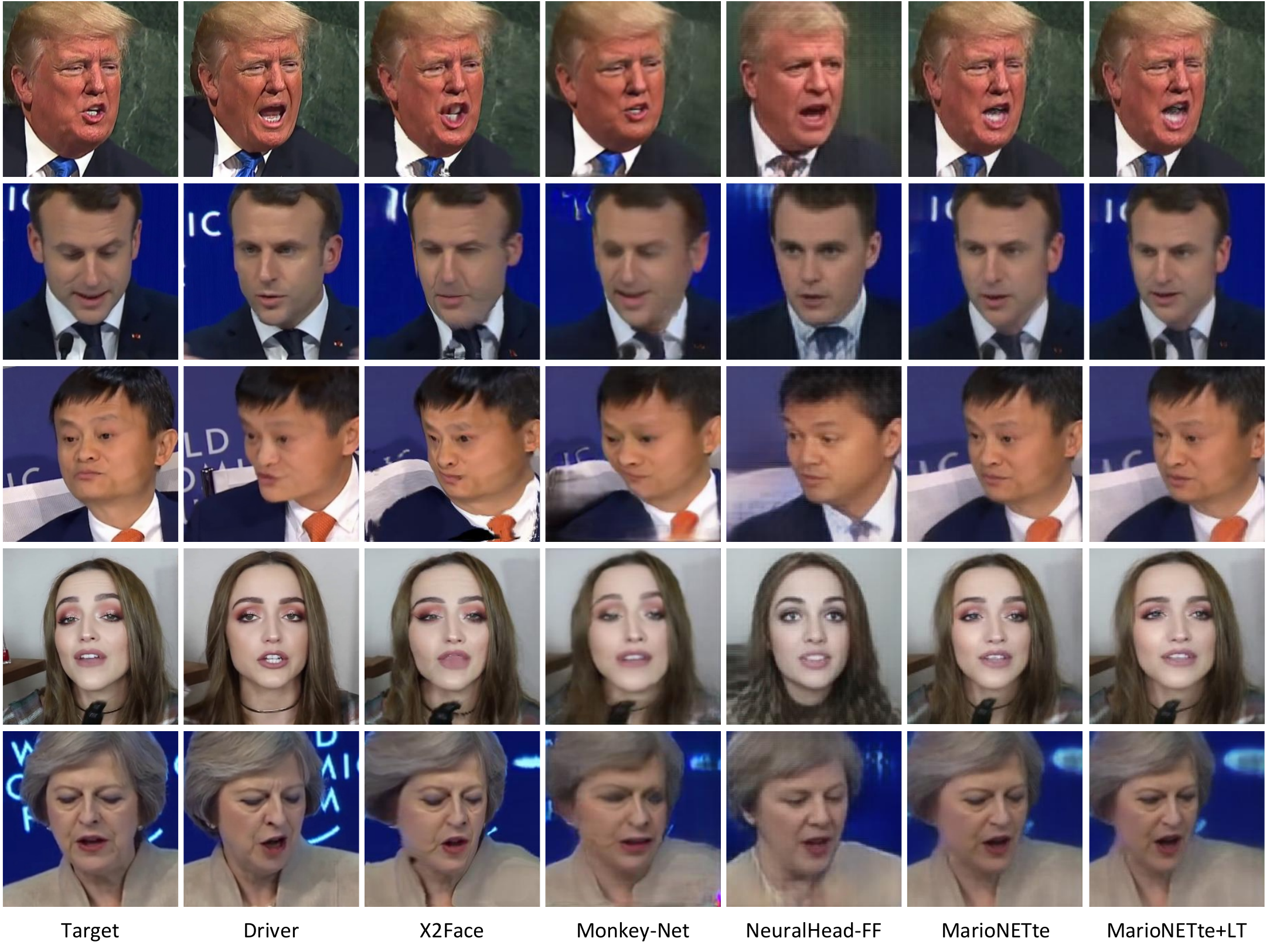}
    \caption{Qualitative results of one-shot self-reenactment setting on CelebV.}
    \label{fig:celebv_same_id_realism_1ref}
\end{figure*}

\begin{figure*}[t]
    \includegraphics[width=0.99\linewidth]{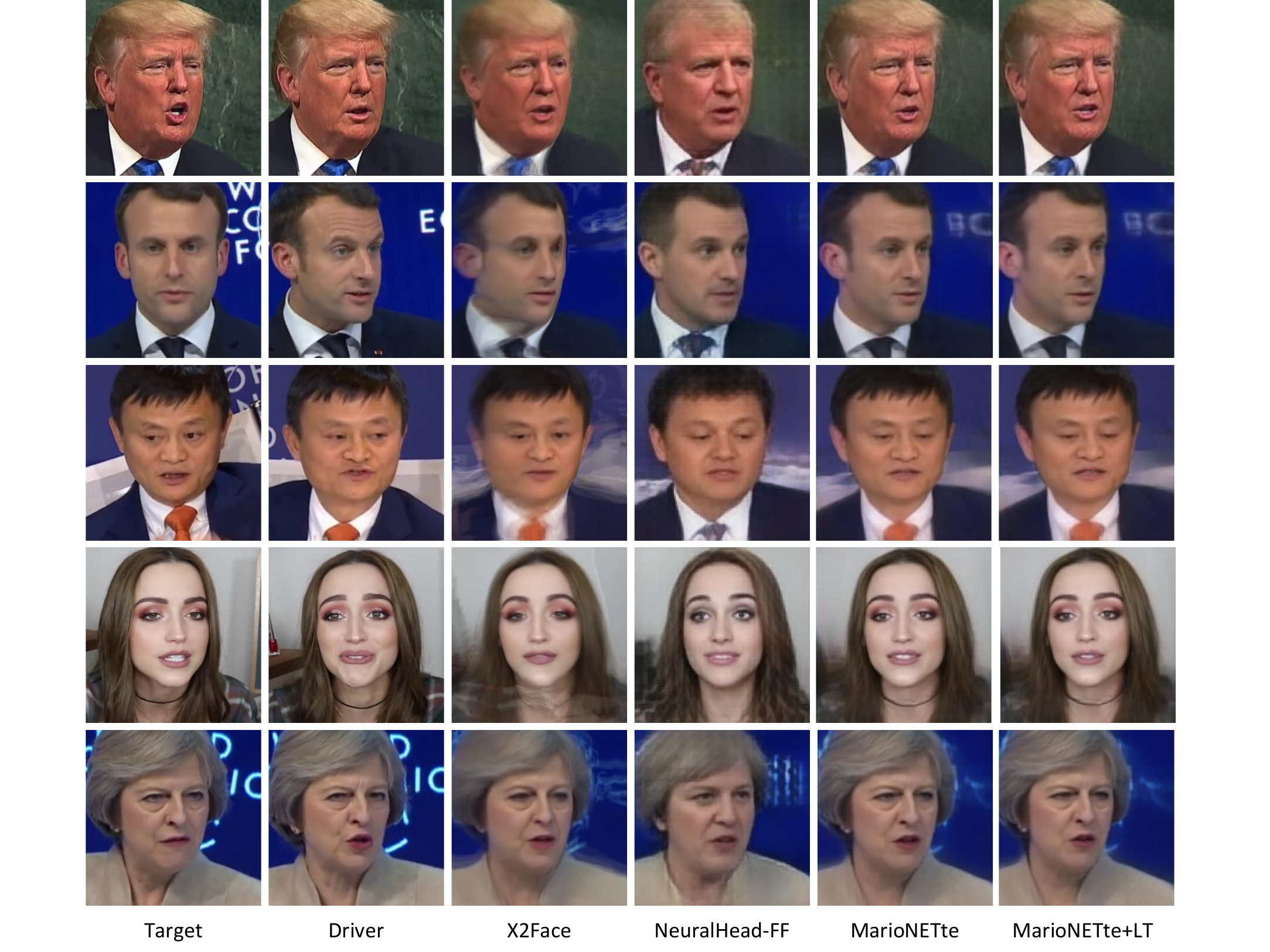}
    \caption{Qualitative results of few-shot self-reenactment setting on CelebV.}
    \label{fig:celebv_same_id_reenact_8ref}
\end{figure*}

\begin{figure*}[t]
    \includegraphics[width=0.99\linewidth]{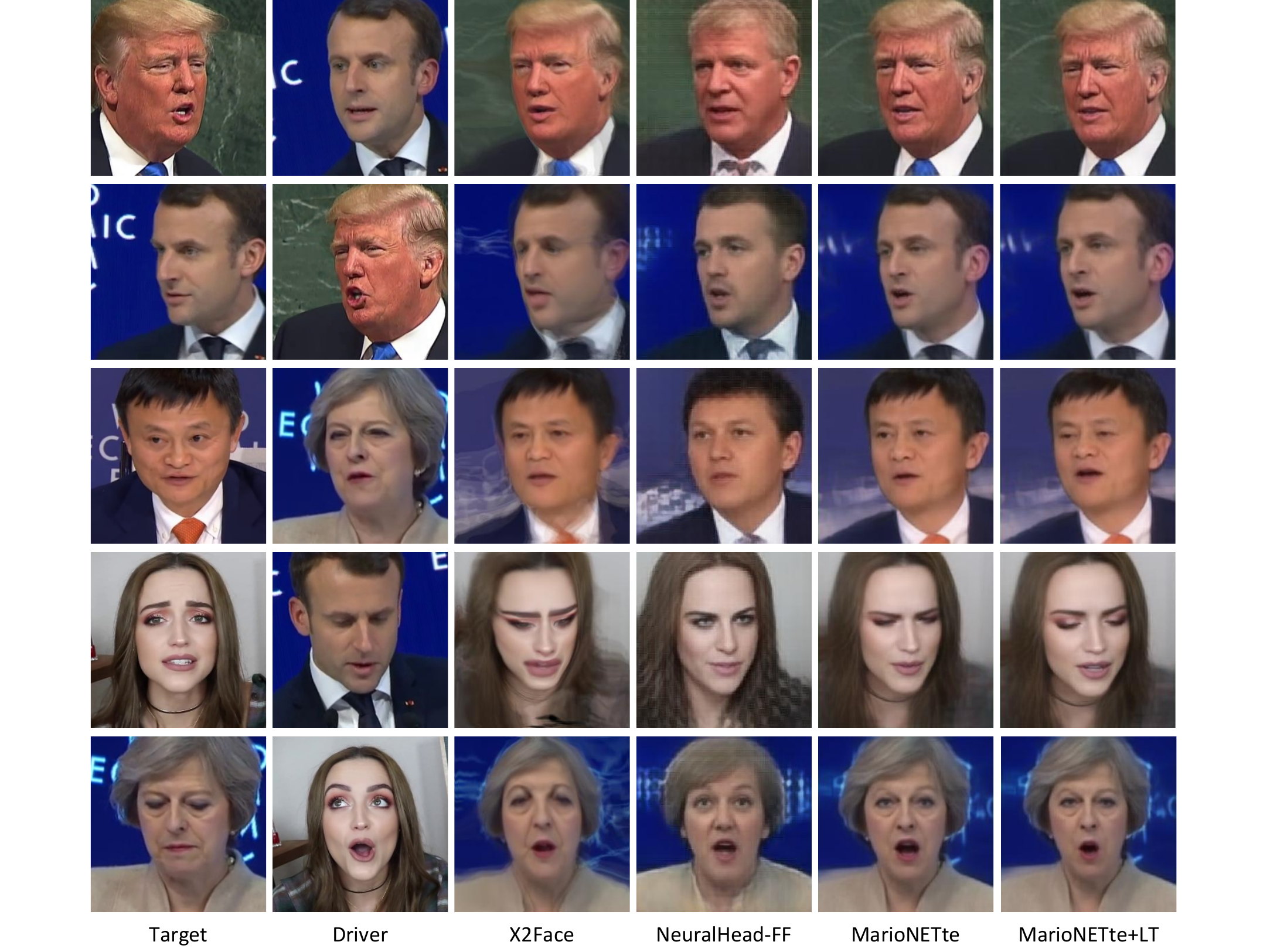}
    \caption{Qualitative results of few-shot reenactment under different identity setting on CelebV.}
    \label{fig:celebv_diff_id_reenact_8ref}
\end{figure*}

Figure~\ref{fig:celebv_same_id_realism_1ref}, Figure~\ref{fig:celebv_same_id_reenact_8ref}, and Figure~\ref{fig:celebv_diff_id_reenact_8ref} depict the qualitative results on the CelebV dataset.
One-shot and few-shot self-reenactment settings of various methods are compared in Figures~\ref{fig:celebv_same_id_realism_1ref} and Figure~\ref{fig:celebv_same_id_reenact_8ref}, respectively.
The results of reenacting a different identity on CelebV under the few-shot setting can be found in Figure~\ref{fig:celebv_diff_id_reenact_8ref}.

\begin{figure*}[t]
    \includegraphics[width=0.99\linewidth]{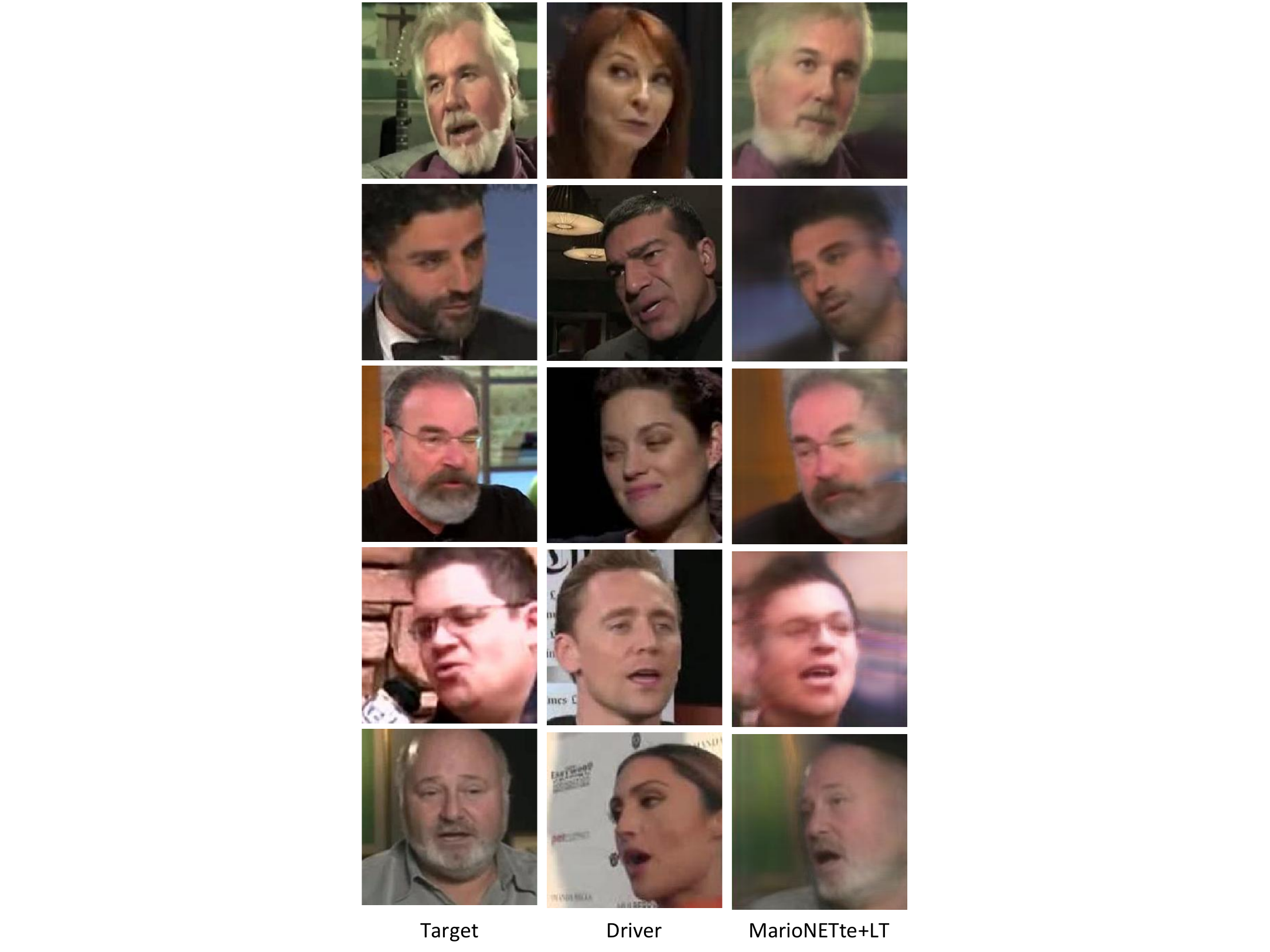}
    \caption{Failure cases generated by MarioNETte+LT while performing one-shot reenactment under different identity setting on VoxCeleb1.}
    \label{fig:marionet_failure}
\end{figure*}

Figure~\ref{fig:marionet_failure} reveals failure cases generated by \textit{MarioNETte+LT} while performing a one-shot reenactment under different identity setting on VoxCeleb1.
Large pose difference between the driver and the target seems to be the main reason for the failures.

\end{appendices}

\end{document}